\documentclass{article} % For LaTeX2e
\usepackage[preprint]{neurips_data_2023}
\usepackage{natbib}
\setcitestyle{authoryear}
\usepackage{tablefootnote}
\usepackage[utf8]{inputenc} % allow utf-8 input
\usepackage[T1]{fontenc}    % use 8-bit T1 fonts
\usepackage[backref=page]{hyperref}
\usepackage[table,xcdraw]{xcolor} % coloured tables
\definecolor{bluecite}{HTML}{0875b7}
\usepackage{float}
\hypersetup{
  colorlinks=true,
  citecolor=bluecite,
  linkcolor=magenta
}
\usepackage{url}            % simple URL typesetting
\usepackage{booktabs}       % professional-quality tables
\usepackage{amsfonts}       % blackboard math symbols
\usepackage{nicefrac}       % compact symbols for 1/2, etc.
\usepackage{microtype}      % microtypography
\usepackage{pifont}
\newcommand{\cmark}{\ding{51}}
\newcommand{\xmark}{\ding{55}}
\usepackage{graphicx}
\graphicspath{ {./figures/} }
\usepackage{adjustbox}

\usepackage[compatibility=false]{caption}
\usepackage{subcaption}
\usepackage{multirow}
\usepackage{bm}

\usepackage[newfloat,frozencache,cachedir=minted-cache]{minted}
\usepackage{tcolorbox}

\SetupFloatingEnvironment{listing}{name=Code Snippet} 

\tcbuselibrary{minted,breakable,xparse,skins}

\definecolor{bg}{gray}{0.95}
\DeclareTCBListing{mintedbox}{O{}m!O{}}{%
  breakable=true,
  listing engine=minted,
  listing only,
  minted language=#2,
  minted style=default,
  minted options={%
    gobble=0,
    breaklines=true,
    breakafter=,,
    fontsize=\tiny,
    numbersep=8pt,
    #1},
  boxsep=0pt,
  left skip=0pt,
  right skip=0pt,
  left=5pt,
  right=5pt,
  top=3pt,
  bottom=3pt,
  arc=5pt,
  leftrule=0pt,
  rightrule=0pt,
  bottomrule=2pt,
  toprule=2pt,
  colback=bg,
  colframe=orange!70,
  enhanced,
  overlay={%
    \begin{tcbclipinterior}
    \fill[orange!20!white] (frame.south west) rectangle ([xshift=1pt]frame.north west);
    \end{tcbclipinterior}},
  #3}

\title{Off-the-Grid MARL: Datasets with Baselines for Offline Multi-Agent Reinforcement Learning}

\author{%
    Claude Formanek \\
    InstaDeep \& University of Cape Town \\
    Cape Town, South Africa \\
    \texttt{c.formanek@instadeep.com} \\
    \And
    Asad Jeewa \\
    University of KwaZulu-Natal\\
    Durban, South Africa \\
    \texttt{jeewaa1@ukzn.ac.za} \\
    \And
    Jonathan Shock \\
    University of Cape Town, NiTheCS \& INRS \\ 
    Cape Town, South Africa, \& Montreal, Canada \\
    \texttt{jonathan.shock@uct.ac.za} \\
    \And
    Arnu Pretorius \\
    InstaDeep\\
    Cape Town, South Africa \\
    \texttt{a.pretorius@instadeep.com} \\
}

\begin{document}

\maketitle

%%%%%%%%%%%%%%%%%%%%%%%%%%%%%%%%%%%%%%%%%%%%%%%%%%%%%%%%%%%%
\begin{abstract}
Being able to harness the power of large datasets for developing cooperative multi-agent controllers promises to unlock enormous value for real-world applications. 
Many important industrial systems are multi-agent in nature and are difficult to model using bespoke simulators. 
However, in industry, distributed processes can often be recorded during operation, and large quantities of demonstrative data stored.
Offline multi-agent reinforcement learning (MARL) provides a promising paradigm for building effective decentralised controllers from such datasets. 
However, offline MARL is still in its infancy and therefore lacks standardised benchmark datasets and baselines typically found in more mature subfields of reinforcement learning (RL). 
These deficiencies make it difficult for the community to sensibly measure progress. 
In this work, we aim to fill this gap by releasing \emph{off-the-grid MARL (OG-MARL)}: a growing repository of high-quality datasets with baselines for cooperative offline MARL research.
Our datasets provide settings that are characteristic of real-world systems, including complex environment dynamics, heterogeneous agents, non-stationarity, many agents, partial observability, suboptimality, sparse rewards and demonstrated coordination.
For each setting, we provide a range of different dataset types (e.g. \texttt{Good}, \texttt{Medium}, \texttt{Poor}, and \texttt{Replay}) and profile the composition of experiences for each dataset. We hope that OG-MARL
will serve the community as a reliable source of datasets and help drive progress, while also providing an accessible entry point for researchers new to the field. The repository for OG-MARL can be found at: \href{https://github.com/instadeepai/og-marl}{https://github.com/instadeepai/og-marl}
\end{abstract}

% , while also providing an accessible entry point for researchers new to the field.
\section{Introduction}
 RL algorithms typically require extensive online interactions with an environment to be able to learn robust policies \citep{yu2018towards}. This limits the extent to which previously-recorded experience may be leveraged for RL applications, forcing practitioners to instead rely heavily on optimised environment simulators that are able to run quickly and in parallel on modern compute hardware. 

In a simulation, it is not atypical to be able to generate years of operating behaviour of a specific system \citep{openai2019dota5, Vinyals2019alphastar}.
However, achieving this level of online data generation throughput in real-world systems, where a realistic simulator is not readily available, can be challenging or near impossible.
More recently, the field of offline RL has offered a solution to this challenge by bridging the gap between RL and supervised learning. In offline RL, the aim is to develop algorithms that are able to leverage large existing datasets of sequential decision-making to learn optimal control strategies that can be deployed online~\citep{levine2020offline}. 
Many researchers believe that offline RL could help unlock the full potential of RL when applied to the real world, where success has been limited~\citep{dulac2021challenges}.

Although the field of offline RL has experienced a surge in research interest in recent years \citep{prudencio2023offlinerl}, the focus on offline approaches specific to the multi-agent setting has remained relatively neglected, despite the fact that many real-world problems are naturally formulated as multi-agent systems (e.g. traffic management \citep{zhang2019cityflow}, a fleet of ride-sharing vehicles \citep{sykora2020marvin}, a network of trains \citep{mohanty2020flatland} or electricity grid management \citep{khattar2022citylearn}). Moreover, systems that require multiple agents (programmed and/or human) to execute coordinated strategies to perform optimally, arguably have a higher barrier to entry when it comes to creating bespoke simulators to model their online operating behaviour. 

Offline RL research in the single agent setting has benefited greatly from publicly available datasets and benchmarks such as D4RL~\citep{fu2020d4rl} and RL Unplugged~\citep{gulcehre2020rl}. Without such offerings in the multi-agent setting to help standardise research efforts and evaluation, it remains challenging to gauge the state of the field and reproduce results from previous work. Ultimately, to develop new ideas that drive the field forward, standardised sets of tasks and baselines are required.

In this paper, we present OG-MARL, a rich set of datasets specifically curated for cooperative offline MARL. We generated diverse datasets on a range of popular cooperative MARL environments. For each environment, we provide different types of behaviour resulting in \textit{Good}, \textit{Medium} and \textit{Poor} datasets as well as \textit{Replay} datasets (a mixture of the previous three). We developed and applied a quality assurance methodology to validate our datasets to ensure that they contain a diverse spread of experiences. Together with our datasets, we provide initial baseline results using state-of-the-art offline MARL algorithms. 

The OG-MARL code and datasets are publicly available through our website.\footnote{\url{https://sites.google.com/view/og-marl}} Additionally, we invite the community to contribute their own datasets to the growing repository on OG-MARL and use our website as a platform for storing and distributing datasets for the benefit of the research community. We hope the lessons contained in our methodology for generating and validating datasets help future researchers to produce high-quality offline MARL datasets and help drive progress. 

\begin{figure}
\begin{minipage}{0.68\textwidth}
    \centering
    \includegraphics[width=\textwidth]{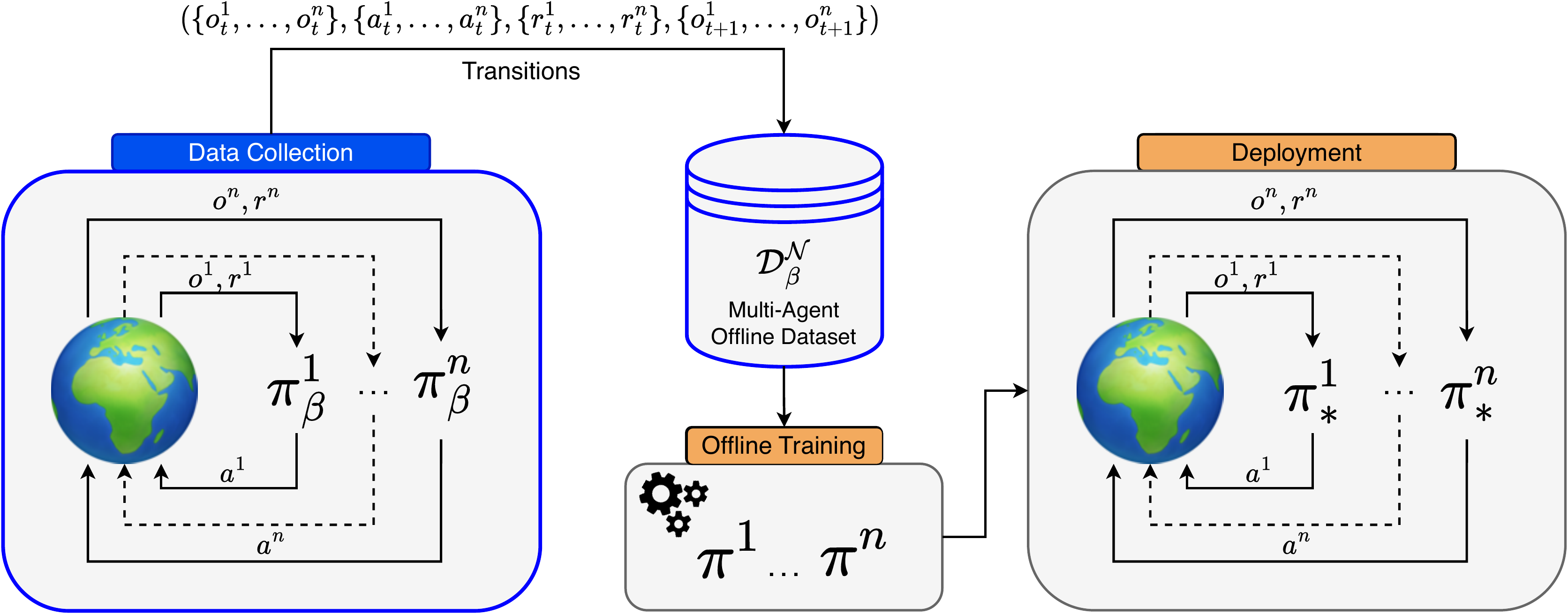}
    \caption{\textbf{Top}: an illustration of offline MARL. Behaviour policies collect experiences and store them in an offline dataset. New policies are trained from the offline data without any online environment interactions. At the end of training, the policies are deployed in the environment. \textbf{Right}: a code snippet demonstrating how to record new datasets, as well as load existing ones, using OG-MARL.}
    \label{fig:og-marl}
\end{minipage}
\hspace{0.5em}
\begin{minipage}{0.29\textwidth}
\centering
\begin{mintedbox}[fontsize=\tiny]{python}
from og_marl import SMAC
from og_marl import QMIX
from og_marl import OfflineLogger

# Instantiate environment
env = SMAC("3m")

# Wrap env in offline logger
env = OfflineLogger(env)

# Make multi-agent system
system = QMIX(env)

# Collect data
system.run_online()

# Load dataset
dataset = env.get_dataset("Good")

# Train offline
system.run_offline(dataset)
\end{mintedbox}
\end{minipage}
\begin{minipage}{0.05\textwidth}
\end{minipage}
\end{figure}
\section{Related Work}

\textbf{Datasets. }
In the single-agent RL setting, D4RL~\citep{fu2020d4rl} and RL~Unplugged~\citep{gulcehre2020rl} have been important contributions, providing a comprehensive set of offline datasets for benchmarking offline RL algorithms. 
While not originally included, D4RL was later extended by~\cite{lu2022challenges} to incorporate datasets with pixel-based observations, which they highlight as a notable deficiency of other datasets. The ease of access to high-quality datasets provided by D4RL and RL Unplugged has enabled the field of offline RL to make rapid progress over the past years \citep{kostrikov2021offline, ghasemipour2022so, nakamoto2023cal}. However, these repositories lack datasets for MARL, which we believe, alongside additional technical difficulties such as large joint action spaces~\citep{yang2021believe}, has resulted in slower progress in the field.

\textbf{Offline Multi-Agent Reinforcement Learning. }
To date, there has been a limited number of papers published on cooperative offline MARL, resulting in benchmarks, datasets and algorithms that do not adhere to any unified standard, making comparisons between works difficult. In brief,
\citet{zhang2021finitesample} carried out an in-depth theoretical analysis of finite-sample offline MARL. \cite{jiang2021marldecentralised} proposed a decentralised multi-agent version of the popular offline RL algorithm BCQ \citep{fujimoto2019bcq} and evaluated it on their own datasets of a multi-agent version of MuJoCo (MAMuJoCo) \citep{peng2021facmac}. \citet{yang2021believe} highlighted how extrapolation error accumulates rapidly in the number of agents and propose a new method they call \textit{Implicit Constraint Q-Learning} (ICQ) to address this. The authors evaluate their method on their own datasets collected using the popular \textit{StarCraft Mulit-Agent Challenge} (SMAC) \citep{samvelyan2019smac}. \citet{pan2022plan} showed that \textit{Conservative Q-Learning} (CQL) \citep{kumar2020cql}, a very successful offline RL method, does not transfer well to the multi-agent setting since the multi-agent policy gradients are prone to uncoordinated local optima. To overcome this, the authors proposed a zeroth-order optimization method to better optimize the conservative value functions, and evaluate their method on their own datasets of a handful of SMAC scenarios, the two agent HalfCheetah scenario from MAMuJoCo and some simple Multi Particle Environments (MPE) \citep{lowe2017maddpg}. \citet{meng2021madt} propose a \textit{multi-agent decision transformer} (MADT) architecture, which builds on the \textit{decision transformer} (DT) \citep{chen2021decision}, and demonstrated how it can be used for offline pre-training and online fine-tuning in MARL by evaluating their method on their own SMAC datasets. \cite{barde2023modelbased} explored a model-based approach for offline MARL and evaluated their method on MAMuJoCo. 

\textbf{Datasets and baselines for Offline MARL. } In all of the aforementioned works, the authors generate their own datasets for their experiments and provide only a limited amount of information about the composition of their datasets (e.g. spread of episode returns and/or visualisations of the behaviour policy). Furthermore, each paper proposes a novel algorithm and typically compares their proposal to a set of baselines specifically implemented for their work. The lack of commonly shared benchmark datasets and baselines among previous papers has made it difficult to compare the relative strengths and weaknesses of these contributions and is one of the key motivations for our work.

Finally, we note works that have already made use of the pre-release version of OG-MARL. \cite{formanek2023reduce} investigated selective ``reincarnation'' in the multi-agent setting and \cite{zhu2023madiff} explored using diffusion models to learn policies in offline MARL. Both these works made use of OG-MARL datasets for their experiments, which allows for easier reproducibility and more sound comparison with future work using OG-MARL.
\section{Preliminaries}
\textbf{Multi-Agent Reinforcement Learning.} There are three main formulations of MARL tasks: competitive, cooperative and mixed. The focus of this work is on the cooperative setting. Cooperative MARL can be formulated as a \textit{decentralised partially observable Markov decision process} (Dec-POMDP)~\citep{bernstein2002decpomdp}. A Dec-POMDP consists of a tuple $\mathcal{M}=(\mathcal{N}, \mathcal{S}, \{\mathcal{A}^i\}, \{\mathcal{O}^i\}$, $P$, $E$, $\rho_0$, $r$, $\gamma)$, where $\mathcal{N}\equiv \{1,\dots,n\}$ is the set of $n$ agents in the system and $s\in \mathcal{S}$ describes the full state of the system. The initial state distribution is given by $\rho_0$. Each agent $i\in \mathcal{N}$ receives only partial information from the environment in the form of a local observation $o^i_t$, given according to an emission function $E(o_t|s_t, i)$. At each timestep $t$, each agent chooses an action $a^i_t\in\mathcal{A}^i$ to form a joint action $\mathbf{a}_t\in\mathcal{A} \equiv \prod^N_i \mathcal{A}^i$. Due to partial observability, each agent typically maintains an observation history $o^i_{0:t}=(o^i_0,\dots, o^i_t)$, or implicit memory, on which it conditions its policy $\mu^i(a_{t}^i|o^i_{0:t})$, when choosing an action. The environment then transitions to a new state in response to the joint action selected in the current state, according to the state transition function $P(s_{t+1}|s_t, \mathbf{a}_t)$ and provides a shared scalar reward to each agent according to a reward function $r(s,a):\mathcal{S}\times \mathcal{A} \rightarrow \mathbb{R}$. We define an agent's return as its discounted cumulative rewards over the $T$ episode timesteps, $G=\sum_{t=0}^T \gamma^t r_t$, where $\gamma \in (0, 1]$ is the discount factor. The goal of MARL in a Dec-POMDP is to find a joint policy $(\pi^1, \dots, \pi^n) \equiv \mathbf{\pi}$ such that the return of each agent $i$, following $\pi^i$, is maximised with respect to the other agents’ policies, $\pi^{-i} \equiv (\pi \backslash \pi^i)$. That is, we aim to find $\pi$ such that
$\forall i: \pi^i \in {\arg\max}_{\hat{\pi}^i} \mathbb{E}\left[G \mid \hat{\pi}^i, \pi^{-i}\right]$

\textbf{Offline Reinforcement Learning.} An offline RL algorithm is trained on a static, previously collected dataset $\mathcal{D}_\beta$  of transitions $(o_t, a_t, r_t, o_{t+1})$ from some (potentially unknown) behaviour policy $\pi_\beta$, without any further online interactions. There are several well-known challenges in the offline RL setting which have been explored, predominantly in the single-agent literature. The primary issues are related to different manifestations of data distribution mismatch between the offline data and the induced online data. \cite{levine2020offline} provide a detailed survey of the problems and solutions in offline RL.

\textbf{Offline Multi-Agent Reinforcement Learning.} In the multi-agent setting, offline MARL algorithms are designed to learn an optimal \textit{joint} policy $(\pi^1, \dots, \pi^n) \equiv \mathbf{\pi}$, from a static dataset $\mathcal{D}_\beta^\mathcal{N}$ of previously collected multi-agent transitions $(\{o^1_t,\dots,o^n_t\},\{a^1_t,\dots,a^n_t\},\{r_t^1,\dots,r_t^n\},\{o^1_{t+1},\dots,o^n_{t+1}\})$, generated by a set of interacting behaviour policies $(\pi^1_\beta, \dots,~\pi^n_\beta) \equiv \mathbf{\pi_\beta}$.
\section{Task Properties}
In order to design an offline MARL benchmark which is maximally useful to the community, we carefully considered the properties that the environments and datasets in our benchmark should satisfy. A major drawback in most prior work has been the limited diversity in the tasks that the algorithms were evaluated on. \cite{meng2021madt} for example only evaluated their algorithm on SMAC datasets and \cite{jiang2021marldecentralised} only evaluated on MAMuJoCo datasets. This makes it difficult to draw strong conclusions about the generalisability of offline MARL algorithms. Moreover, these environments fail to test the algorithms along dimensions which may be important for real-world applications. In this section, we outline the properties we believe are important for evaluating offline MARL algorithms.

\textbf{Centralised and Independent Training. } The environments supported in OG-MARL are designed to test algorithms that use decentralised execution, i.e.\ at execution time, agents need to choose actions based on their local observation histories only. However, during training, centralisation (i.e.\ sharing information between agents) is permissible, although not required. \textit{Centralised training with decentralised execution} (CTDE) \citep{kraemer2016multi} is one of the most popular MARL paradigms and is well-suited for many real-world applications. Being able to test both centralised and independent training algorithms is important because it has been shown that neither paradigm is consistently better than the other \citep{lyu2021contrasting}. As such, both types of algorithms can be evaluated using OG-MARL datasets and we also provide baselines for both centralised and independent training.

\textbf{Different types of Behaviour Policies.} We generated datasets with several different types of behaviour policies including policies trained using online MARL with fully independent learners (e.g. independent DQN and independent TD3), as well as CTDE algorithms (e.g. QMIX and MATD3).  Furthermore, some datasets generated with CTDE algorithms used a state-based critic while others used a joint-observation critic. It was important for us to consider both of these critic setups as they are known to result in qualitatively different policies \citep{lyu2022deeper}. More specific details of which algorithms were used to generate which datasets can be found in \autoref{tab:all_environments} in the appendix.

\textbf{Partial Information.} It is common for agents to receive only local information about their environment, especially in real-world systems that rely on decentralised components. Thus, some of the environments in OG-MARL test an algorithm's ability to leverage agents' \textit{memory} in order to choose optimal actions based only on partial information from local observations. This is in contrast to settings such as MAMuJoCo where prior methods \citep{jiang2021marldecentralised, pan2022plan} achieved reasonable results without instilling agents with any form of memory.

\textbf{Different Observation Modalities.} In the real world, agent observations come in many different forms. For example, observations may be in the form of a feature vector or a matrix representing a pixel-based visual observation. \cite{lu2022challenges} highlighted that prior single-agent offline RL datasets failed to test algorithms on high-dimensional pixel-based observations. OG-MARL tests algorithms on a diverse set of observation modalities, including feature vectors and pixel matrices of different sizes.

\textbf{Continuous and Discrete Action Spaces.} The actions an agent is expected to take can be either discrete or continuous across a diverse range of applications. Moreover, continuous action spaces can often be more challenging for offline MARL algorithms as the larger action spaces make them more prone to extrapolation errors, due to out-of-distribution actions \cite{}. OG-MARL supports a range of environments with both discrete and continuous actions.

\textbf{Homogeneous and Heterogeneous Agents.} Real-world systems can either be homogeneous or heterogeneous in terms of the types of agents that comprise the system. In a homogeneous system, it may be significantly simpler to train a single policy and copy it to all agents in the system. On the other hand, in a heterogenous system, where agents may have significantly different roles and responsibilities, this approach is unlikely to succeed. OG-MARL provides datasets from environments that represent both homogeneous and heterogeneous systems.

\textbf{Number of Agents.} Practical MARL systems may have a large number of agents. Most prior works to date have evaluated their algorithms on environments with typically fewer than 8 agents \citep{pan2022plan, yang2021believe, jiang2021marldecentralised}. In OG-MARL, we provide datasets with between 2 and 27 agents, to better evaluate \textit{large-scale} offline MARL (see \autoref{tab:all_environments}). 

\textbf{Sparse Rewards.} Sparse rewards are challenging in the single-agent setting, but in the multi-agent setting, it can be even more challenging due to the multi-agent credit assignment problem \citep{zhou2020learning}. Prior works focused exclusively on dense reward settings \citep{pan2022plan,yang2021believe}. To overcome this, OG-MARL also provides datasets with sparse rewards. 

\textbf{Team and Individual Rewards.} Some environments have team rewards while others can have an additional local reward component. Team rewards exacerbate the multi-agent credit assignment problem, and having a local reward component can help mitigate this. However, local rewards may result in sub-optimality, where agents behave too greedily with respect to their local reward and as a result jeopardize achieving the overall team objective. OG-MARL includes tasks to test algorithms along both of these dimensions.

\textbf{Procedurally Generated and Stochastic Environments.} Some popular MARL benchmark environments are known to be highly deterministic \citep{ellis2022smacv2}. This limits the extent to which the generalisation capabilities of algorithms can be evaluated. Procedurally generated environments have proved to be a useful tool for evaluating generalisation in single-agent RL \citep{cobbe2020}. In order to better evaluate generalisation in offline MARL, OG-MARL includes stochastic tasks that make use of procedural generation.

\textbf{Realistic Multi-Agent Domains.} Almost all prior offline MARL works have evaluated their algorithms exclusively on game-like environments such as StarCraft~\citep{yang2021believe} and particle simulators~\citep{pan2022plan}. Although a large subset of open research questions may still be readily investigated in such simulated environments, we argue that in order for offline MARL to become more practically relevant, benchmarks in the research community should begin to closer reflect real-world problems of interest. Therefore, in addition to common game-like benchmark environments, OG-MARL also supports environments which simulate more real-world like problems including energy management and control~\citep{vazquezcanteli2020citylearn,wang2021voltage}. While there remains a large gap between these environments and truly real-world settings, it is a step in the right direction to keep pushing the field forward and enable useful contributions in the development of new algorithms and improving our understanding of key difficulties and failure modes.

\textbf{Human Behaviour Policies.} The current standard practice in the offline MARL literature is to use policies trained using RL as the behaviour policies for offline MARL datasets. However, for real-world applications, behaviour policies are likely to be non-RL policies such as human operators or hand-crafted controllers. In order to encourage the offline MARL community to move beyond RL behaviour policies, we provide a dataset of humans playing the \textit{Knights, Archers and Zombies} game from PettingZoo. It is our hope that this contribution will catalyse more research on offline MARL from human-generated data.

\textbf{Competitive Scenarios.} Competitive offline MARL is an under researched area with only a handful of existing works in the field \citep{cui2022provably}. Moreover, all of the existing works have almost exclusively focused on tabular two-player zero-sum Markov games \citep{cui2021minimax,zhong2022pessimistic}. In order to encourage the offline MARL research community to make advances in the competitive offline setting, we provide datasets on two popular competitive \textit{MPE} environments. Furthermore, the offline data recorder in OG-MARL can readily be used by researchers to generate their own competitive offline datasets on novel environments.
\section{Environments}

\begin{figure}
\centering
    \begin{subfigure}[t]{0.175395\textwidth}
        \centering
        \includegraphics[width=\textwidth]{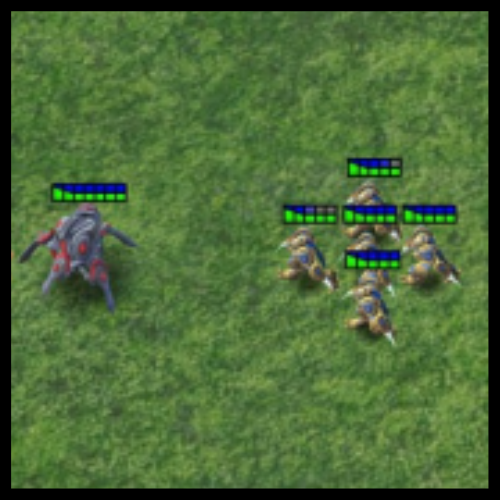}
        \caption{SMAC v1 \& v2}
        \label{fig:smac}
        \centering
        \includegraphics[width=\textwidth]{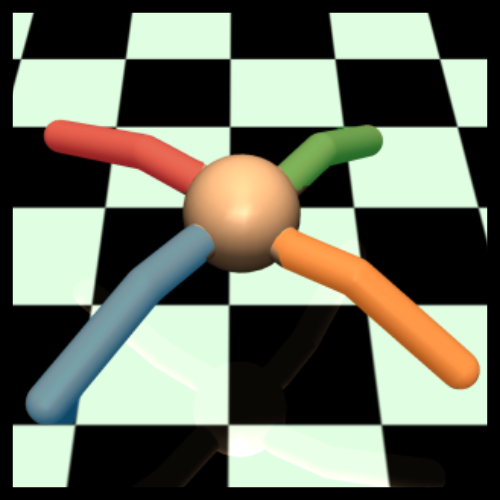}
        \caption{MAMuJoCo}
        \label{fig:mamujoco}
    \end{subfigure}
    \begin{subfigure}[t]{0.175395\textwidth}
        \centering
        \includegraphics[width=\textwidth]{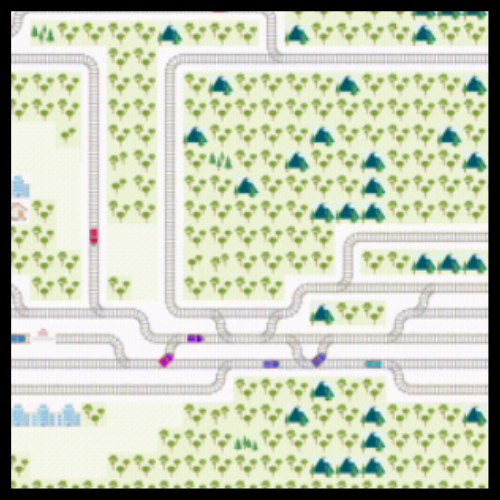}
        \caption{Flatland}
        \label{fig:flatland}
        \centering
        \includegraphics[width=\textwidth]{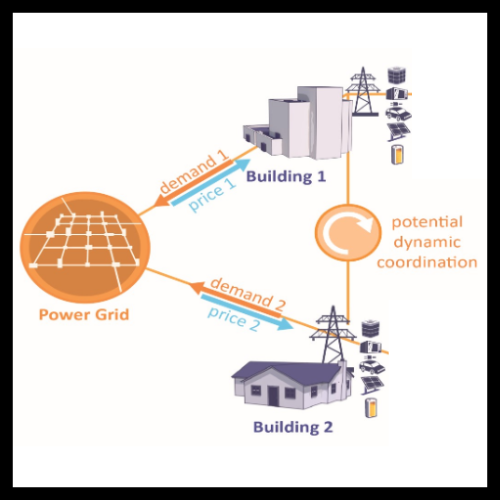}
        \caption{CityLearn}
        \label{fig:citylearn}
    \end{subfigure}
    \begin{subfigure}[t]{0.175395\textwidth}
        \centering
        \includegraphics[width=\textwidth]{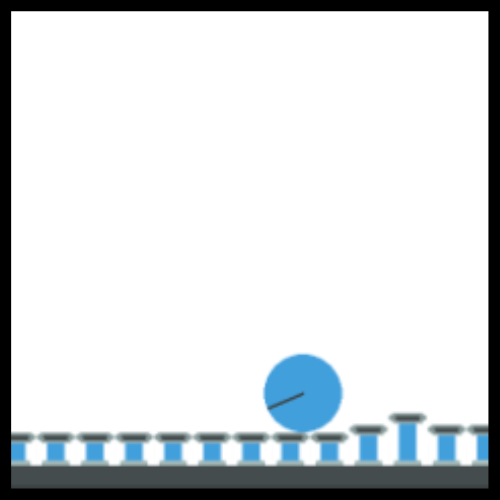}
        \caption{Pistonball}
        \label{fig:pistonball}
        \centering
        \includegraphics[width=\textwidth]{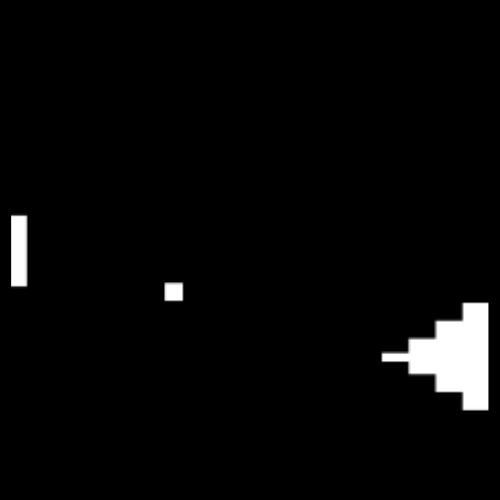}
        \caption{Co-op Pong}
        \label{fig:coop_pong}
    \end{subfigure}
    \begin{subfigure}[t]{0.175395\textwidth}
        \centering
        \includegraphics[width=\textwidth]{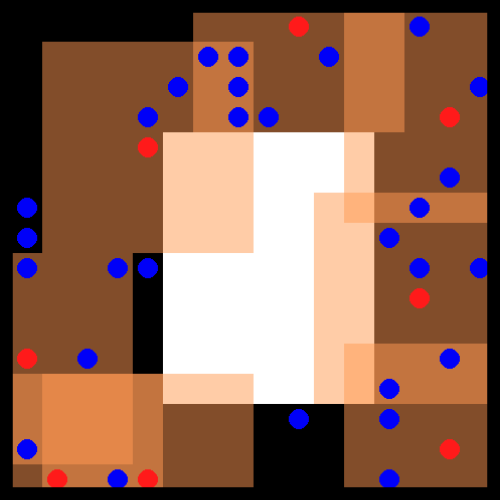}
        \caption{Pursuit}
        \label{fig:pursuit}
        \centering
        \includegraphics[width=\textwidth]{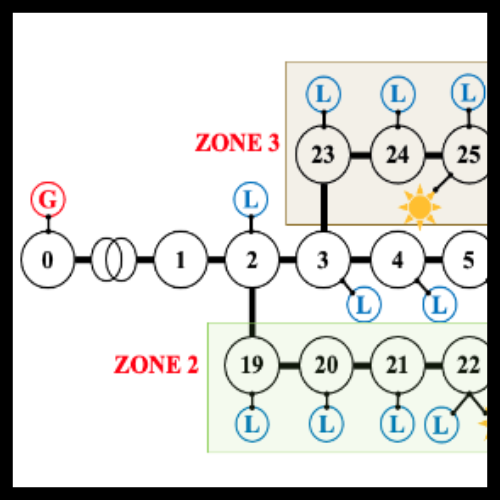}
        \caption{Voltage Control}
        \label{fig:voltage}
    \end{subfigure}
\caption{MARL environments for which we provide datasets in OG-MARL.}
    \label{fig:envs}
\end{figure}

\textbf{SMAC v1} \textit{(hetero- and homogeneous agents, local observations)}. SMAC is the most popular cooperative offline MARL environment used in the literature\citep{gorsane2022emarl}. SMAC focuses on the micromanagement challenge in StarCraft 2 where each unit is controlled by an independent agent that must learn to cooperate and coordinate based on local (partial) observations. SMAC played an important role in moving the MARL research community beyond grid-world problems and has also been very popular in the offline MARL literature \citep{yang2021believe, meng2021madt, pan2022plan}. Thus, it was important for OG-MARL to support a range of SMAC scenarios.

\textbf{SMAC v2} \textit{(procedural generation, local observations)}. Recently some deficiencies in SMAC have been brought to light. Most importantly, SMAC is highly deterministic, and agents can therefore learn to \textit{memorise} the best policy by conditioning on the environment timestep only. To address this, SMACv2 \citep{ellis2022smacv2} was recently released and includes non-deterministic scenarios, thus providing a more challenging benchmark for MARL algorithms. In OG-MARL, we publicly release the first set of SMACv2 datasets.

\textbf{MAMuJoCo} \textit{(hetero- and homogeneous agents, continuous actions)}. The MuJoCo environment \citep{todorov2012mujoco} has been an important benchmark that helped drive research in continuous control. More recently, MuJoCo has been adapted for the multi-agent setting by introducing independent agents that control different subsets of the whole MuJoCo robot (MAMuJoCo) \citep{peng2021facmac}. MAMuJoCo is an important benchmark because there are a limited number of continuous action space environments available to the MARL research community. MAMuJoCo has also been widely adopted in the offline MARL literature \citep{jiang2021marldecentralised, pan2022plan}. Thus, in OG-MARL we provide the largest openly available collection of offline datasets on scenarios in MAMuJoCo (\cite{pan2022plan}, for example, only provided a single dataset on 2-Agent HalfCheetah).

\textbf{PettingZoo} \textit{(pixel observations, discrete and continuous actions)}. OpenAI's Gym \citep{brockman2016openai} has been widely used as a benchmark for single agent RL. PettingZoo is a gym-like environment-suite for MARL \citep{terry2021pettingzoo} and provides a diverse collection of environments. In OG-MARL, we provide a general-purpose environment wrapper which can be used to generate new datasets for any PettingZoo environment. Additionally, we provide initial datasets on three PettingZoo environments including \textit{PistonBall}, \textit{Co-op Pong} and \textit{Pursuit}~\citep{gupta2017cooperative}. We chose these environments because they have visual (pixel-based) observations of varying sizes; an important dimension along which prior works have failed to evaluate their algorithms.

\textbf{Flatland} \textit{(real-world problem, procedural generation, sparse local rewards)}. The train scheduling problem is a real-world challenge with significant practical relevance. Flatland~\citep{mohanty2020flatland} is a simplified 2D simulation of the train scheduling problem that is an appealing benchmark for cooperative MARL for several reasons. Firstly, it randomly generates a new train track layout and timetable at the start of each episode, thus testing the generalisation capabilities of MARL algorithms to a greater degree than many other environments. Secondly, Flatland has a very sparse and noisy reward signal, as agents only receive a reward on the final timestep of the episode. Finally, agents have access to a local reward component. These properties make the Flatland environment a novel, challenging and realistic benchmark for offline MARL. 

\textbf{Voltage Control and CityLearn} \textit{(real-world problem, continuous actions)}. Energy management \citep{yu2021energy} is another appealing real-world application for MARL, especially given the large potential efficiency gains and corresponding positive effects on climate change that could be had~\citep{rolnick2022climate}. As such, we provide datasets for two challenging MARL environments related to energy management. Firstly, we provide datasets for the \textit{Active Voltage Control on Power Distribution Networks} environment \citep{wang2021voltage}. Secondly, we provide datasets for the CityLearn environment \citep{vazquezcanteli2020citylearn} where the goal is to develop agents for distributed energy resource management and demand response between a network of buildings with batteries and photovoltaics.

\textbf{Knights, Archers \& Zombies} \textit{(human behaviour policies)}. In \textit{Knights, Archers and Zombies}~\citep{terry2021pettingzoo} zombies walk from the top border of the screen down to the bottom border in unpredictable paths. The agents controlled are a knight and an archer which can each move around and attack the zombies. The game ends when all agents die (collide with a zombie) or a zombie reaches the bottom screen border. We collected experience of several different combinations of human players. The players where given no instruction on how to play the game and where allowed to play a maximum of 20 episodes.

\textbf{MPE} \textit{(competitive)}. MPE is a popular suite of multi-agent environments, first introduced by \citep{lowe2017maddpg}. We provide datasets of two competitive scenarios, namely \textit{Simple Adversary} and \textit{Simple Push} \citep{terry2021pettingzoo}. Additional information about these competitive datasets is included in the appendix.

\section{Datasets}

To generate the transitions in the datasets, we recorded environment interactions of partially trained online algorithms, as has been common in prior works for both single-agent \citep{gulcehre2020rl} and multi-agent settings \citep{yang2021believe,pan2022plan}. For discrete action environments, we used QMIX~\citep{rashid2018qmix} and independent DQN \cite{} and for continuous action environments, we used independent TD3 \citep{fujimoto2018td3} and MATD3~\citep{lowe2017maddpg, ackermann2019matd3}. Additional details about how each dataset was generated are included in \autoref{section:appendix_datasets}.

\begin{figure}
    \centering
    \begin{subfigure}[]{0.32\textwidth}
        \includegraphics[width=\textwidth]{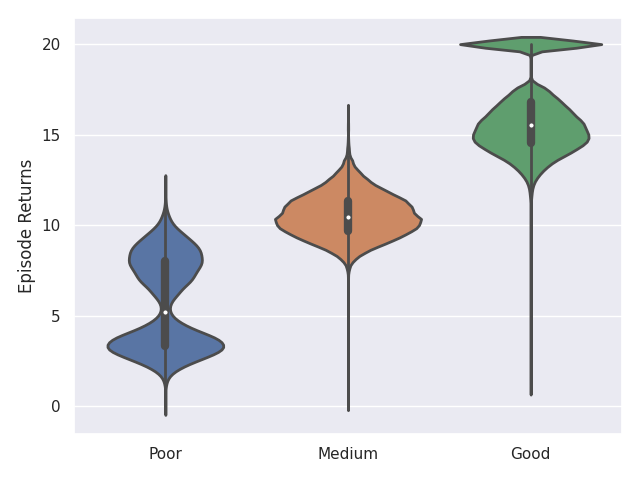}
        \caption{SMAC 27m\_vs\_30m}
        \label{fig:violin_27m_vs_30m}
    \end{subfigure}
    \begin{subfigure}[]{0.32\textwidth}
        \includegraphics[width=\textwidth]{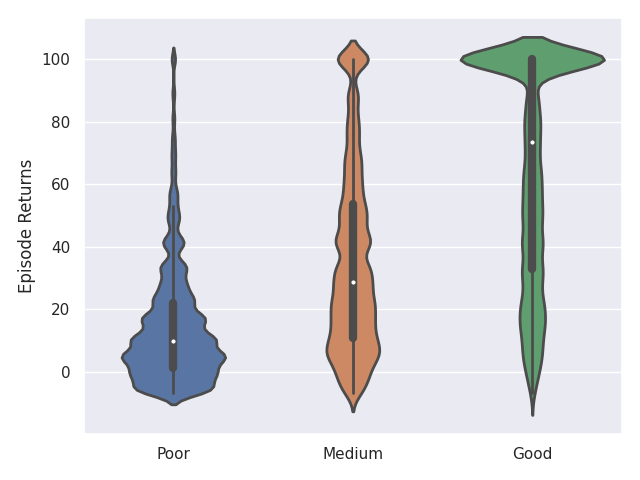}
        \caption{PettingZoo Co-op Pong}
        \label{fig:violin_coop_pong}
    \end{subfigure}
    \begin{subfigure}[]{0.32\textwidth}
        \includegraphics[width=\textwidth]{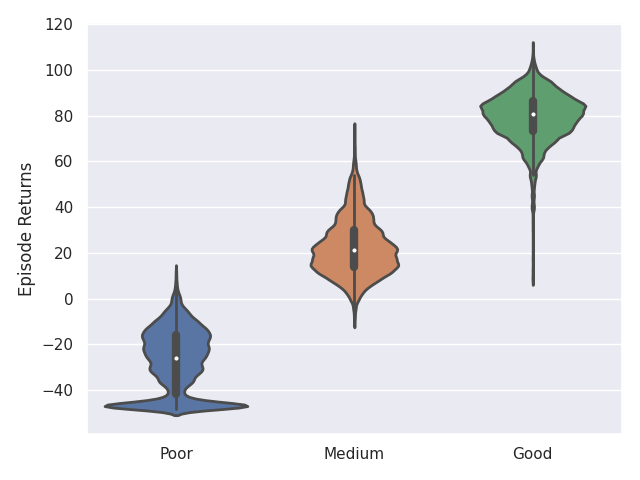}
        \caption{PettingZoo Pursuit}
        \label{fig:violin_pursuit}
    \end{subfigure}
    \caption{Violin plots of the probability distribution of episode returns for selected datasets in OG-MARL. In blue the \texttt{Poor} datasets, in orange the \texttt{Medium} datasets and in green the \texttt{Good} datasets. Wider sections of the violin plot represent a higher probability of sampling a trajectory with a given episode return, while the thinner sections correspond to a lower probability. The violin plots also include the median, interquartile range and min/max episode return for the datasets.}
    \label{fig:violin}
\end{figure}

\textbf{Diverse Data Distributions.} It is well known from the single-agent offline RL literature that the quality of experience in offline datasets can play a large role in the final performance of offline RL algorithms \citep{fu2020d4rl}. In OG-MARL, we include a range of dataset distributions including \texttt{Good}, \texttt{Medium}, \texttt{Poor} and \texttt{Replay} datasets in order to benchmark offline MARL algorithms on a range of different dataset qualities. The dataset types are characterised by the quality of the joint policy that generated the trajectories in the dataset, which is the same approach taken in previous works \citep{meng2021madt, yang2021believe, pan2022plan}. To ensure that all of our datasets have sufficient coverage of the state and action spaces, while also containing minimal repetition i.e.\ not being too narrowly focused around a single strategy, we used 3 independently trained joint policies to generate each dataset, and additionally added a small amount of exploration noise to the policies. The boundaries for the different categories were assigned independently for each environment and were related to the maximum attainable return in the environment. Additional details about how the different datasets were curated can be found in \autoref{section:appendix_datasets}.

\textbf{Statistical characterisation of datasets.} It is common in both the single-agent and multi-agent offline RL literature for researchers to curate offline datasets by unrolling episodes using an RL policy that was trained to a desired \textit{mean} episode return. However, authors seldom report the distribution of episode returns induced by the policy. Reporting only the mean episode return of the behaviour policy can be  misleading~\citep{agarwal2021deep}. To address this, we provide violin plots to visualise the distribution of expected episode returns. A violin plot is a powerful tool for visualising numerical distributions as they visualise the density of the distribution as well as several summary statistics such as  the minimum, maximum and interquartile range of the data. These properties make the violin plot very useful for understanding the distribution of episode returns in the offline datasets, assisting with interpreting offline MARL results. \autoref{fig:violin} provides a sample of the violin plots for different scenarios (the remainder of the plots can be found in the appendix). In each figure, the difference in shape and position of the three violins illustrates the difference in the datasets with respect to the expected episode return. Additionally, we provide a table with the mean and standard deviation of the episode returns for each of the datasets in \autoref{tab:all_datasets}, similar to \cite{meng2021madt}.

\section{Baselines}\label{sec:baselines}

\begin{table}
\centering
\small
\caption{Results on the \textit{Pursuit} and \textit{Co-op Pong} datasets. The mean episode return with one standard deviation across all seeds is given. In each row the best mean episode return is in bold.}
\vspace{1em}
\begin{tabular}{lclllll}
\textbf{Scenario}        & \textbf{Dataset} & \multicolumn{1}{c}{\textbf{BC}} & \multicolumn{1}{c}{\textbf{QMIX}} & \multicolumn{1}{c}{\textbf{QMIX+BCQ}} & \multicolumn{1}{c}{\textbf{QMIX+CQL}} & \multicolumn{1}{c}{\textbf{MAICQ}} \\ \hline
\multirow{3}{*}{Co-op Pong}    & Good             & 31.2$\pm$3.5                    & 0.6$\pm$3.5                       & 1.9$\pm$1.1                           & \textbf{90.0$\pm$4.7}                 & 75.4$\pm$3.9                       \\
                         & Medium           & 21.6$\pm$4.8                    & 10.6$\pm$17.6                     & 20.3$\pm$12.2                         & 64.9$\pm$15.0                         & \textbf{84.6$\pm$0.9}              \\
                         & Poor             & 1.0$\pm$0.9                     & 14.4$\pm$16.0                     & 30.2$\pm$20.7                         & 52.7$\pm$8.5                          & \textbf{74.8$\pm$7.8}              \\ \hline
\multirow{3}{*}{Pursuit} & Good             & 78.3$\pm$1.8                    & 6.7$\pm$19.0                      & 66.9$\pm$14.0                         & 54.4$\pm$6.3                          & \textbf{92.7$\pm$3.7}              \\
                         & Medium           & 15.0$\pm$1.6                    & -24.4$\pm$20.2                    & 16.6$\pm$10.7                         & 20.6$\pm$10.3                         & \textbf{35.3$\pm$3.0}              \\
                         & Poor             & -18.5$\pm$1.6                   & -43.7$\pm$5.6                     & \textbf{-0.7$\pm$4.0}                 & -19.6$\pm$3.3                         & -4.1$\pm$0.7                       \\ \midrule 
\bottomrule
\end{tabular}
\label{tab:pettingzoo_results}
\end{table}

 In this section, we present the initial baselines that we provide with OG-MARL. This serves two purposes: \textit{i)} to validate the quality of our datasets and \textit{ii)} to enable the community to use these initial results for development and performance comparisons in future work. In the main text, we present results on two PettingZoo environments (\textit{Pursuit} and \textit{Co-op Pong}), since these environments and their corresponding datasets are a novel benchmark for offline MARL. Furthermore, it is the first set of environments with pixel-based observations to be used to evaluate offline MARL algorithms. We include all additional baseline results in \autoref{tab:all_discrete_results} and \autoref{tab:all_cont_results}.

 \textbf{Baseline Algorithms.} State-of-the-art algorithms were implemented from seminal offline MARL work. For discrete action environments we implemented \textit{Behaviour Cloning} (BC), QMIX \citep{rashid2018qmix}, QMIX with \textit{Batch Constrained Q-Learning} \citep{fujimoto2019bcq} (QMIX+BCQ), QMIX with \textit{Conservative Q-Learning} \citep{kumar2020cql} (QMIX+CQL) and MAICQ \citep{yang2021believe}. For continuous action environments, Behaviour Cloning (BC), Independent TD3 (ITD3), ITD3 with \textit{Behaviour Cloning} regularisation \citep{fujimoto2021td3bc} (ITD3+BC), ITD3 with \textit{Conservative Q-Learning} (ITD3+CQL) and OMAR \citep{pan2022plan} were implemented. \autoref{sec:appendix_baselines} provides additional implementation details on the baseline algorithms.

 \textbf{Experimental Setup.} On \textit{Pursuit} and \textit{Co-op Pong}, all of the algorithms were trained offline for $50 000$ training steps with a fixed batch size of $32$. At the end of training, we evaluated the performance of the algorithms by unrolling the final joint policy in the environment for $100$ episodes and recording the mean episode return over the episodes. We repeated this procedure for $10$ independent seeds as per the recommendation by \citet{gorsane2022emarl}. We kept the online evaluation budget \citep{kurenkov2022onlineEvalBudget} fixed for all algorithms by only tuning hyper-parameters on \textit{Co-op Pong} and keeping them fixed for \textit{Pursuit}. Controlling for the online evaluation budget is important when comparing offline algorithms because online evaluation may be expensive, slow or dangerous in real-world problems, making online hyper-parameter fine-tuning infeasible. See \autoref{sec:appendix_baselines} for a further discussion on hyper-parameter tuning in OG-MARL.

 \textbf{Results.} In \autoref{tab:pettingzoo_results} we provide the unnormalised mean episode returns for each of the discrete action algorithms on the different datasets for \textit{Pursuit} and \textit{Co-op Pong}. 
 % Results on the other environments and datasets can be found in \autoref{tab:all_discrete_results} and \autoref{tab:all_cont_results}.

\begin{figure}
    \begin{subfigure}{0.3\textwidth}
        \includegraphics[width=\textwidth]{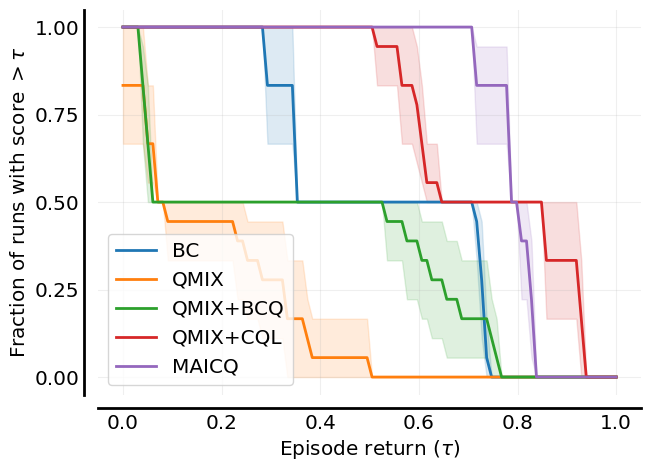}
        \caption{Good}
    \end{subfigure}
    \begin{subfigure}{0.3\textwidth}
        \includegraphics[width=\textwidth]{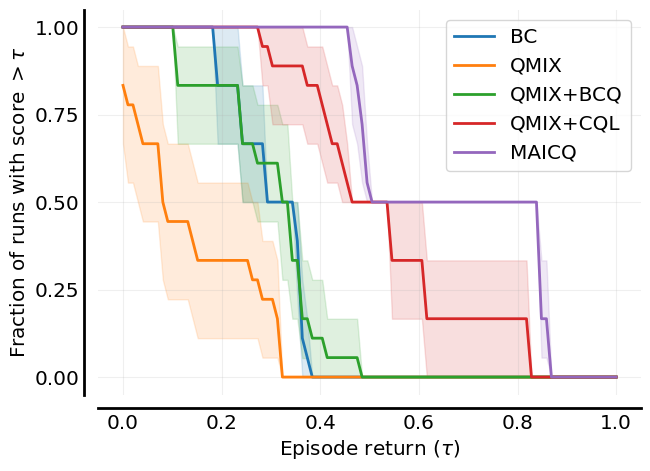}
        \caption{Medium}
    \end{subfigure}
    \begin{subfigure}{0.3\textwidth}
        \includegraphics[width=\textwidth]{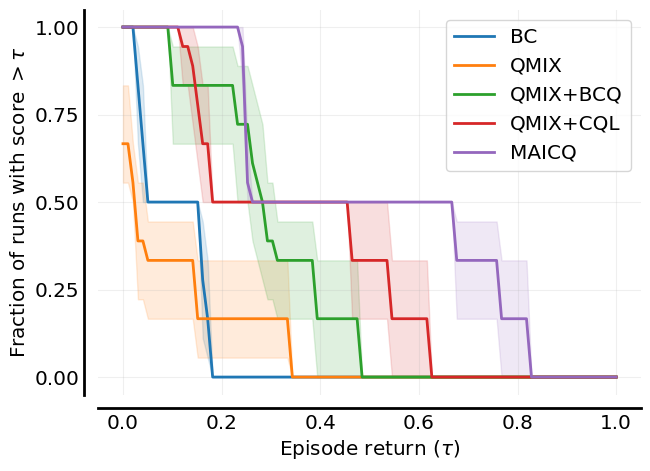}
        \caption{Poor}
    \end{subfigure}
    \caption{Performance profiles \citep{agarwal2021deep} aggregated across all seeds on \textit{Pursuit} and \textit{Co-op Pong}. Shaded regions show pointwise 95\% confidence bands based on percentile bootstrap with stratified sampling.}
    \label{fig:perf_profiles}
\end{figure}

\textbf{Aggregated Results.} In addition to the tabulated results we also provide \textit{aggregated} results as per the recommendation by \cite{gorsane2022emarl}. In \autoref{fig:perf_profiles} we plot the performance profiles \citep{agarwal2021deep} of the discrete action algorithms by aggregating across all seeds and the two environments, \textit{Pursuit} and \textit{Co-op Pong}. To facilitate aggregation across environments, where the possible episode returns can be very different, we adopt the normalisation procedure from \citet{fu2020d4rl}. On the \texttt{Good} datasets, we found that MAICQ and QMIX+CQL both outperformed behaviour cloning (BC). QMIX+BCQ did not outperform BC and vanilla QMIX performed very poorly. On the \texttt{Medium} datasets, MAICQ and QMIX+CQL once again performed the best, significantly outperforming BC. QMIX+BCQ marginally outperformed BC and vanilla QMIX failed. Finally, on the \texttt{Poor} datasets, MAICQ, QMIX+CQL and QMIX+BCQ all outperformed BC but MAICQ was the best by some margin. These results on PettingZoo environments, with pixel observations, further substantiate that MAICQ is the current state-of-the-art offline MARL algorithm in discrete action settings. 

\section{Discussion}

\textbf{Limitations and future work.} \label{sec:limitations}The primary limitation of this work is that it focuses on the cooperative setting. Additionally, the datasets used in OG-MARL were exclusively generated by online MARL policies. Future work could explore the inclusion of datasets from alternate sources, such as hand-designed or human controllers, which may exhibit distinct properties~\citep{fu2020d4rl}. Moreover, an exciting research direction considers the offline RL problem as a sequence modeling task~\citep{chen2021decision, meng2021madt}, and we aim to incorporate such models as additional baselines in OG-MARL in future iterations.

\textbf{Potential Negative Societal Impacts. } While the potential positive impacts of efficient decentralized controllers powered by offline MARL are promising, it is essential to acknowledge and address the potential negative societal impacts \citep{whittlestone2021societal}. Deploying a model trained using offline MARL in real-world applications requires careful consideration of safety measures \citep{gu2022safety, xu2022constraints}. Practitioners should exercise caution to ensure the safe and responsible implementation of such models.

\textbf{Conclusion. } In this work, we highlighted the importance of offline MARL as a research direction for applying RL to real-world problems. We specifically focused on the lack of a standard set of benchmark datasets, which is a significant obstacle to progress. To address this issue, we presented a set of relevant and diverse datasets for offline MARL. We profiled our datasets by visualising the distribution of episode returns in violin plots and tabulated mean and standard deviations. We validated our datasets by providing a set of initial baseline results with state-of-the-art offline MARL algorithms. Finally, we open-sourced all of our software tooling for generating new datasets and provided a website with our code, as well as for hosting and sharing the datasets. It is our hope that the research community will adopt and contribute towards OG-MARL as a framework for offline MARL research and that it helps to drive progress in this nascent field.

\newpage
% Include in deanonymised version.

% \section*{Acknowledgements}
% We would like to thank Callum Rhys Tilbury for his valuable comments on this work.

% TODO Double check bib style for neurips
\bibliographystyle{abbrvnat}
\bibliography{main}

% Checklist
% \newpage
% \input{sections/10-checklist}

% Appendix
\newpage
\appendix
\counterwithin{figure}{section}
\counterwithin{table}{section}

\newpage
\section{Datasheet} \label{sec:appendix_datasheet}

The following is the datasheet \citep{gebru2021datasheets} for Off-the-Grid Multi-Agent Reinforcement Learning (OG-MARL).
The OG-MARL is openly accessible on our GitHub repository and datasets are available for download from our website:

\begin{itemize}
    \item \url{https://sites.google.com/view/og-marl}
    \item \url{https://github.com/instadeepai/og-marl}
\end{itemize}

\subsection{Motivation}

\textbf{For what purpose was the dataset created?} The datasets in OG-MARL were created to facilitate research in offline Multi-Agent Reinforcement Learning (MARL). Offline MARL is a nascent field of machine learning that promises to unlock real-world applications of MARL. However, progress has been hampered by the lack of a standardised, high-quality benchmark datasets. OG-MARL was built to fill this gap and drive progress in the field.

\textbf{Who created the dataset and on behalf of which entity?} OG-MARL was created by Claude Formanek, Asad Jeewa, Jonathan Shock and Arnu Pretorious on behalf of InstaDeep and the University of Cape Town.

\textbf{Who funded the creation of the dataset?} The creation of OG-MARL was funded by InstaDeep.

\subsection{Composition} 

\textbf{What do the instances that comprise the dataset represent?} The various datasets in OG-MARL comprise of environment transitions in popular MARL benchmark environments (e.g. SMAC by \cite{samvelyan2019smac}). The transitions were generated by recording environment interactions between policies trained using online RL. 

\textbf{How many instances are there in total?} Each dataset in OG-MARL has approximately 1 million transitions in it.

\textbf{Does the dataset contain all possible instances or is it a sample of instances from a larger set?} Great care was taken to ensure that the dataset in OG-MARL had good coverage of the state and action space of the environment. It is however, not possible (nor desirable) to guarantee full coverage.

\textbf{What data does each instance consist of?} Each instance consists of a sequence of multi-agent transitions in the environment. A transition is composed of agent observations, actions, rewards and next observations $(\{o^1_t,\dots,o^n_t\},\{a^1_t,\dots,a^n_t\},\{r_t^1,\dots,r_t^n\},\{o^1_{t+1},\dots,o^n_{t+1}\})$.  

\textbf{Is there a label or target associated with each instance?} As we are in the reinforcement learning paradigm, instances do not have \textit{labels}. However, since each instance is a multi-agent transition, they do each have a corresponding reward for each agent, which we use for training.

\textbf{Is any information missing from individual instances?} Everything is included. No data is missing.

\textbf{Are relationships between individual instances made explicit?} Transitions that belong to the same episode can be retrieved together, if desired.

\textbf{Are there recommended data splits?} In offline RL one does not need to split data like in supervised learning. All data can be used for training.

\textbf{Are there any errors, sources of noise, or redundancies in the dataset?} None.

\textbf{Is the dataset self-contained, or does it link to or otherwise rely on external resources?} OG-MARL is completely self-contained. The datasets are stored in a binary format but can be loaded into a dataset loader with the utilities provided in the OG-MARL code.

\textbf{Does the dataset contain data that might be considered confidential?} No.

\textbf{Does the dataset contain data that, if viewed directly, might be offensive, insulting, threatening, or might otherwise cause anxiety?} No.

\textbf{Does the dataset identify any subpopulations?} No.

\textbf{Is it possible to identify individuals, either directly or indirectly from the dataset?} No.

\textbf{Does the dataset contain data that might be considered sensitive in any way?} No.

\subsection{Collection Process}

\textbf{How was the data associated with each instance acquired?} To generate the datasets for OG-MARL we trained online MARL algorithms on a variety of popular MARL benchmark environments and recorded the environment transitions.

\textbf{What mechanisms or procedures were used to collect the data?} We trained our online MARL algorithms on a PC with a GPU (Nvidia RTX 3070) and recorded experiences with a python utility we designed and subsequently open-sourced to the community.

\textbf{If the dataset is a sample from a larger set, what was the sampling strategy?} The different datasets in OG-MARL have different data compositions. We grouped transitions into \texttt{Good}, \texttt{Medium} and \texttt{Poor} according to the return of episode that the transition belonged to.

\textbf{Who was involved in the data collection process?} <Anonymous> and <anonymous>.

\textbf{Over what timeframe was the data collected?} The datasets in the current version of OG-MARL were collected over a period of about 3 months.

\textbf{Were any ethical review processes conducted?} No, since it is believed that none was required.

\textbf{Did you collect the data from the individuals in question directly, or obtain it via third parties or other sources } No.

\textbf{Were the individuals in question notified about the data collection?} Not applicable.

\textbf{Did the individuals in question consent to the collection and use of their data?} Not applicable.

\textbf{If consent was obtained, were the consenting individuals provided with a mechanism to revoke their consent in the future or for certain uses?} Not applicable.

\textbf{Has an analysis of the potential impact of the dataset and its use on data subjects been conducted?} Not applicable.

\subsection{Preprocessing/Cleaning/Labeling}

\textbf{Was any preprocessing/cleaning/labeling of the data done?} Transitions were grouped into short continuous sequences to allow for training recurrent policies.

\textbf{Was the “raw” data saved in addition to the preprocessed/cleaned/labeled data?} Individual transitions can be loaded instead of sequences.

\textbf{Is the software that was used to preprocess/clean/label the data available?} Yes, in the OG-MARL repository.

\subsection{Uses}

\textbf{Has the dataset been used for any tasks already?} Yes. \cite{zhu2023madiff} and \cite{formanek2023reduce} both used OG-MARL.

\textbf{Is there a repository that links to any or all papers or systems that use the dataset?} There is a repository hosted by InstaDeep linked above.

\textbf{What (other) tasks could the dataset be used for?} OG-MARL could be used for any kind of sequential decision-making research.

\textbf{Is there anything about the composition of the dataset or the way it was collected and preprocessed/cleaned/labeled that might impact future uses? } Loading entire episodes of transitions needs to be made easier in future releases.

\textbf{Are there tasks for which the dataset should not be used?} The data in OG-MARL was generated on simplified environments and does not necessarily generalise to the real world.

\subsection{Distribution}

\textbf{Will the dataset be distributed to third parties outside of the entity on behalf of which the dataset was created?} Yes, OG-MARL is publicly available on the internet.

\textbf{How will the dataset will be distributed?} OG-MARL datasets are hosted in an S3 bucket but can easily be accessed via our publicly open website or by running the download scripts in the OG-MARL code.

\textbf{When will the dataset be distributed?} The datasets were released in February of 2023.

\textbf{Will the dataset be distributed under a copyright or other intellectual property (IP) license, and/or under applicable terms of use (ToU)?} We have applied the \textit{CC BY-NC-SA} dataset licence to OG-MARL.

\textbf{Have any third parties imposed IP-based or other restrictions on the data associated with the instances?} No.

\textbf{Do any export controls or other regulatory restrictions apply to the dataset or to individual instances?} No.

\subsection{Maintenance}

\textbf{Who will be supporting/hosting/maintaining the dataset?} Claude Formanek will be responsible for maintaining the datasets on behalf of InstaDeep, who will also be financially supporting the hosting of the datasets.

\textbf{How can the owner/curator/manager of the dataset be contacted?} Via email, c.formanek@instadeep.com.

\textbf{Is there an erratum?} Versioning and changes are tracked on the OG-MARL repository.

\textbf{Will the dataset be updated?} OG-MARL is a growing collection of offline MARL datasets. The creator and the wider community will be adding new datasets over time.

\textbf{If the dataset relates to people, are there applicable limits on the retention of the data associated with the instances?} Not applicable.

\textbf{Will older versions of the dataset continue to be supported/hosted/maintained?} Yes.

\textbf{If others want to extend/augment/build on/contribute to the dataset, is there a mechanism for them to do so?} Yes, please open a pull request on the OG-MARL repository.

\newpage
\section{Additional Environment Information} \label{section:appendix_environments}

In this section we provide additional information of all of the environments supported in OG-MARL. In \autoref{tab:all_environments} we provide an overview of salient environment characteristics, in addition to the algorithm which was used to generate the behaviour policies. In \autoref{tab:environments_links} we provide links to the source of the environments for the reader to refer to for additional information about the environments.

\begin{table}[H]
\centering
\scriptsize
\caption{All supported environments and scenarios in OG-MARL and some of their characteristics.} \label{tab:all_environments}
\vspace{1em}
\begin{adjustbox}{max width=\textwidth}
\begin{tabular}{ ccccccccc }
Environment & Scenario & Agents & Actions & Observations & Reward & Agent Types & Behaviour & Online Perf.\\
\hline
\multirow{2}{5em}{SMAC} & 3m & 3 & \multirow{7}{*}{Discrete} & \multirow{7}{*}{Vector} & \multirow{7}{*}{Dense} & Homog & \multirow{7}{*}{QMIX} & 16.1 \\  
& 8m & 8 &  &  &  & Homog & & 16.2 \\
& 2s3z & 5 &  &  &  & Heterog & & 18.2\\
& 5m\_vs\_6m & 5 &  &  &  & Homog & & 16.6\\
& 27m\_vs\_30m & 27 &  &  &  & Homog & & 16.0\\
& 3s5z\_vs\_3s6z & 8 &  &  &  & Heterog & & 17.0\\
& 2c\_vs\_64zg & 2 &  &  &  & Homog & & 18.0\\
\hline
\multirow{3}{5em}{MAMuJoCo} & 2-HalfCheetah & 2 & \multirow{3}{*}{Continuous} & \multirow{3}{*}{Vector} & \multirow{3}{*}{Dense} & Heterog & \multirow{3}{*}{MATD3} & 6924 \\
& 2-Ant & 2 &  &  &  & Homog & & 2621 \\
& 4-Ant & 4 &  &  &  & Homog & & 2769 \\
\hline
\multirow{3}{5em}{PettingZoo} & Pursuit & 8 & Discrete & \multirow{3}{*}{Pixels} & \multirow{3}{*}{Dense} & Homog & QMIX & 79.5\\
& Co-op Pong & 2 & Discrete &  &  & Heterog & IDQN & 65.1\\
& Pistonball & 15 & Continuous &  &  & Homog & MATD3 & 84.6\\
\hline
\multirow{2}{5em}{Flatland} & 3 Trains & 3 & \multirow{2}{*}{Discrete} & \multirow{2}{*}{Vector} & \multirow{2}{*}{Dense} & \multirow{2}{*}{Homog} & \multirow{2}{*}{IDQN} & -5.1\\
& 5 Trains & 5 &  &  &  &  & & -5.9\\
\hline
\multirow{2}{5em}{SMAC v2} & terran\_5\_vs\_5 & 5 & \multirow{3}{*}{Discrete} & \multirow{3}{*}{Vector} & \multirow{3}{*}{Dense} & \multirow{3}{*}{Hetrog} & \multirow{3}{*}{QMIX} & 17.0\\
& zerg\_5\_vs\_5 & 5 &  &  &  &  & & 15.2\\
& terran\_10\_vs\_10 & 10 &  &  &  &  & & 16.9\\
\hline
\multirow{1}{*}{CityLearn} & 2022\_all\_phases & 17 & \multirow{1}{*}{Continuous} & \multirow{1}{*}{Vector} & \multirow{1}{*}{Dense} & \multirow{1}{*}{Homog} & ITD3 & -6421 \\
\hline
\multirow{1}{*}{Voltage Control} & case33\_3min\_final & 6 & \multirow{1}{*}{Continuous} & \multirow{1}{*}{Vector} & \multirow{1}{*}{Dense} & \multirow{1}{*}{Homog} & ITD3 & -12.3\\
\hline
\bottomrule
\end{tabular}
\end{adjustbox}
\end{table} \label{tab:appendix_environments}

\begin{table}[H]
\centering
\scriptsize
\caption{All environments with links to their sources.}
\begin{tabular}{ll}
\textbf{Environment} & \textbf{Website}                                          \\ \hline
SMAC v1              & https://github.com/oxwhirl/smac                           \\
SMAC v2              & https://github.com/oxwhirl/smacv2                         \\
PettingZoo           & https://pettingzoo.farama.org/                            \\
Flatland             & https://flatland.aicrowd.com/intro.html                   \\
MAMuJoCo             & https://github.com/schroederdewitt/multiagent\_mujoco     \\
CityLearn            & https://github.com/intelligent-environments-lab/CityLearn \\
Voltage Control      & https://github.com/Future-Power-Networks/MAPDN    \\       
\hline
\bottomrule
\end{tabular}
\label{tab:environments_links}
\end{table}

\newpage
\section{Additional Information on Datasets} \label{section:appendix_datasets}
In this section, we provide additional information about the datasets in OG-MARL. In \autoref{tab:all_datasets} we give the mean episode return with standard deviation for all datasets in OG-MARL.

\begin{table}[H]
\centering
\scriptsize
\caption{Table of the mean episode return with the standard deviation for all datasets in OG-MARL.}
\vspace{1em}
\begin{tabular}{lllcc}
\textbf{Environment}             & \multicolumn{1}{c}{\textbf{Scenario}} & \multicolumn{1}{c}{\textbf{Dataset}} & \textbf{Mean Episode Return ($\pm$ Std)} & \textbf{Number of Sequences} \\ \hline
\multirow{21}{*}{SMAC}           & \multirow{3}{*}{3m}                   & Good                                 & 16.0$\pm$6.1                             & 120569                       \\
                                 &                                       & Medium                               & 10.0$\pm$6.0                             & 120004                       \\
                                 &                                       & Poor                                 & 4.8$\pm$2.3                              & 118447                       \\ \cline{2-5} 
                                 & \multirow{3}{*}{8m}                   & Good                                 & 16.3$\pm$4.4                             & 111873                       \\
                                 &                                       & Medium                               & 10.3$\pm$3.4                             & 120845                       \\
                                 &                                       & Poor                                 & 5.3$\pm$0.6                              & 109515                       \\ \cline{2-5} 
                                 & \multirow{3}{*}{5m\_vs\_6m}           & Good                                 & 16.6$\pm$4.7                             & 112779                       \\
                                 &                                       & Medium                               & 12.8$\pm$5.1                             & 117594                       \\
                                 &                                       & Poor                                 & 7.7$\pm$1.5                              & 110031                       \\ \cline{2-5} 
                                 & \multirow{3}{*}{2s3z}                 & Good                                 & 18.2$\pm$2.9                             & 107900                       \\
                                 &                                       & Medium                               & 12.8$\pm$3.1                             & 107640                       \\
                                 &                                       & Poor                                 & 6.8$\pm$2.1                              & 101197                       \\ \cline{2-5} 
                                 & \multirow{3}{*}{3s5z\_vs3s6z}         & Good                                 & 17.0$\pm$3.3                             & 101335                       \\
                                 &                                       & Medium                               & 11.0$\pm$1.7                             & 107873                       \\
                                 &                                       & Poor                                 & 5.7$\pm$2.3                              & 107475                       \\ \cline{2-5} 
                                 & \multirow{3}{*}{2c\_vs\_64zg}         & Good                                 & 18.0$\pm$2.2                             & 108270                       \\
                                 &                                       & Medium                               & 13.1$\pm$2.0                             & 111199                       \\
                                 &                                       & Poor                                 & 9.9$\pm$1.6                              & 115370                       \\ \cline{2-5} 
                                 & \multirow{3}{*}{27m\_vs\_30m}         & Good                                 & 16.0$\pm$2.1                             & 110271                       \\
                                 &                                       & Medium                               & 10.5$\pm$1.2                             & 113737                       \\
                                 &                                       & Poor                                 & 5.7$\pm$2.5                              & 110845                       \\ \hline
\multirow{9}{*}{MAMuJoCo}        & \multirow{3}{*}{2-HalfCheetah}        & Good                                 & 6924$\pm$1270                            & 100000                       \\
                                 &                                       & Medium                               & 1484$\pm$469                             & 100000                       \\
                                 &                                       & Poor                                 & 400$\pm$333                              & 100000                       \\ \cline{2-5} 
                                 & \multirow{3}{*}{2-Ant}                & Good                                 & 2621$\pm$493                             & 100041                       \\
                                 &                                       & Medium                               & 1099$\pm$264                             & 100109                       \\
                                 &                                       & Poor                                 & 437$\pm$164                              & 99804                        \\ \cline{2-5} 
                                 & \multirow{3}{*}{4-Ant}                & Good                                 & 2769$\pm$270                             & 100170                       \\
                                 &                                       & Medium                               & 1546$\pm$389                             & 100215                       \\
                                 &                                       & Poor                                 & 542$\pm$216                              & 100224                       \\ \hline
\multirow{9}{*}{PettingZoo}      & \multirow{3}{*}{Pursuit}              & Good                                 & 79.5$\pm$10.8                            & 101249                       \\
                                 &                                       & Medium                               & 22.7$\pm$12.4                            & 100087                       \\
                                 &                                       & Poor                                 & -27.3$\pm$14.0                           & 100000                       \\ \cline{2-5} 
                                 & \multirow{3}{*}{Co-op Pong}           & Good                                 & 65.1$\pm$35.6                            & 100687                       \\
                                 &                                       & Medium                               & 35.6$\pm$29.9                            & 101490                       \\
                                 &                                       & Poor                                 & 14.4$\pm$18.7                            & 102277                       \\ \cline{2-5} 
                                 & \multirow{3}{*}{Pistonball}           & Good                                 & 84.6$\pm$17.9                            & 208518                       \\
                                 &                                       & Medium                               & 34.1$\pm$25.6                            & 200142                       \\
                                 &                                       & Poor                                 & 12.0$\pm$22.6                            & 200000                       \\ \hline
\multirow{6}{*}{Flatland}        & \multirow{3}{*}{3 Trains}             & Good                                 & -5.2$\pm$8.0                             & 23000                        \\
                                 &                                       & Medium                               & -16.1$\pm$11.8                           & 19800                        \\
                                 &                                       & Poor                                 & -28.8$\pm$11.8                           & 19200                        \\ \cline{2-5} 
                                 & \multirow{3}{*}{5 Trains}             & Good                                 & -5.9$\pm$8.0                             & 20600                        \\
                                 &                                       & Medium                               & -16.3$\pm$10.2                           & 18000                        \\
                                 &                                       & Poor                                 & -25.5$\pm$10.5                           & 17600                        \\ \hline
\multirow{3}{*}{SMAC v2}         & \multirow{1}{*}{terran\_5\_vs\_5}     & \multicolumn{1}{c}{Replay}           & 10.4$\pm$5.9                             & 97795                        \\ \cline{2-5} 
                                 & \multirow{1}{*}{zerg\_5\_vs\_5}       & \multicolumn{1}{c}{Replay}           & 7.5$\pm$3.6                              & 137776                       \\ \cline{2-5} 
                                 & \multirow{1}{*}{terran\_10\_vs\_10}   & \multicolumn{1}{c}{Replay}           & 11.4$\pm$5.6                             & 75355                        \\ \hline
\multirow{1}{*}{CityLearn}       & \multirow{1}{*}{2022\_all\_phases}    & \multicolumn{1}{c}{Replay}           & -6820.7$\pm$458.4                              & 169068                          \\ \hline
\multirow{1}{*}{Voltage Control} & \multirow{1}{*}{case33\_3min\_final}                  & \multicolumn{1}{c}{Replay}           & -25.1$\pm$22.3                              &  40541                            \\ \hline
\end{tabular}
\label{tab:all_datasets}
\end{table}

\newpage
\subsection{Violin Plots}
In addition to the table with mean episode returns, we also provide violin plots for all datasets in OG-MARL in order to visualise the distribution of episode returns induced by the behaviour policies.

\begin{figure}[H]
    \centering
%    \hfill
    \begin{subfigure}{0.3\textwidth}
        \centering
        \includegraphics[width=\textwidth]{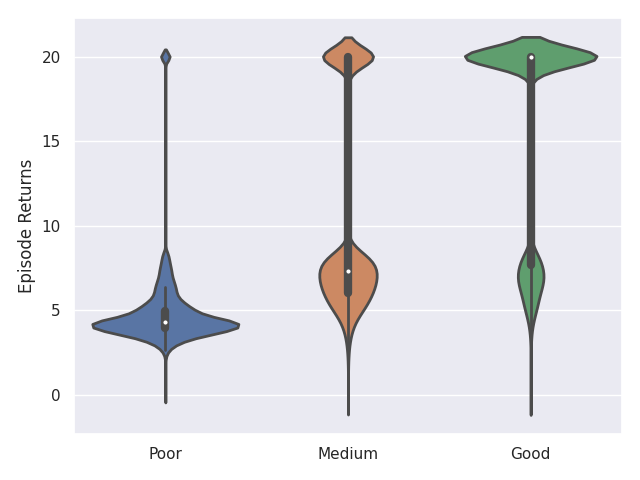}
        \caption{SMAC 3m}
        \label{fig:violin_3m_app}
    \end{subfigure}
    \begin{subfigure}{0.3\textwidth}
        \centering
        \includegraphics[width=\textwidth]{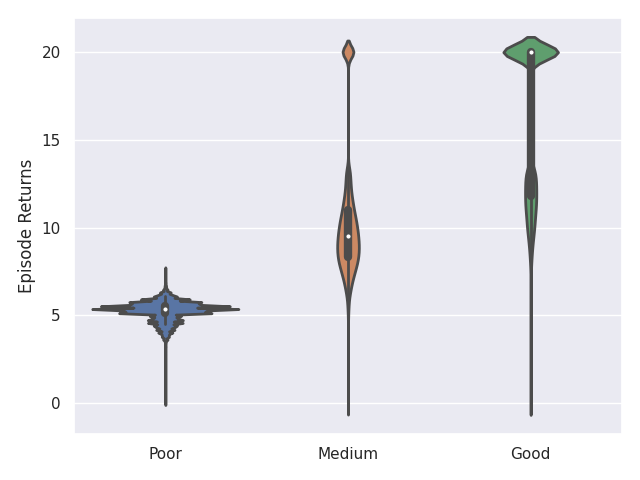}
        \caption{SMAC 8m}
        \label{fig:violin_8m_app}
    \end{subfigure}
%    \hfill
    \begin{subfigure}{0.3\textwidth}
        \centering
        \includegraphics[width=\textwidth]{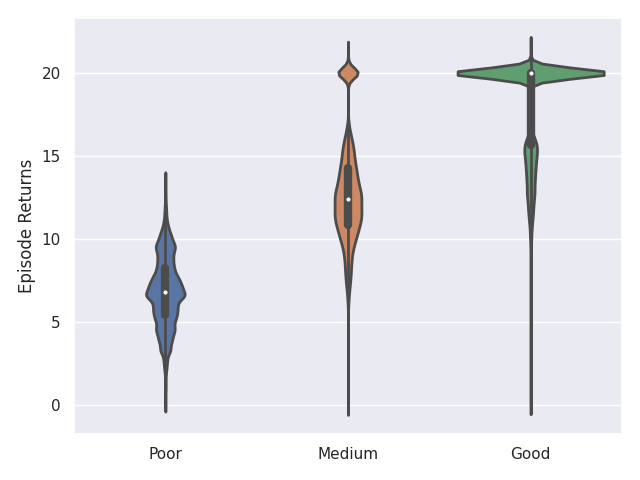}
        \caption{SMAC 2s3z}
        \label{fig:violin_2s3z_app}
    \end{subfigure}
%    \hfill
    \begin{subfigure}{0.3\textwidth}
        \centering
        \includegraphics[width=\textwidth]{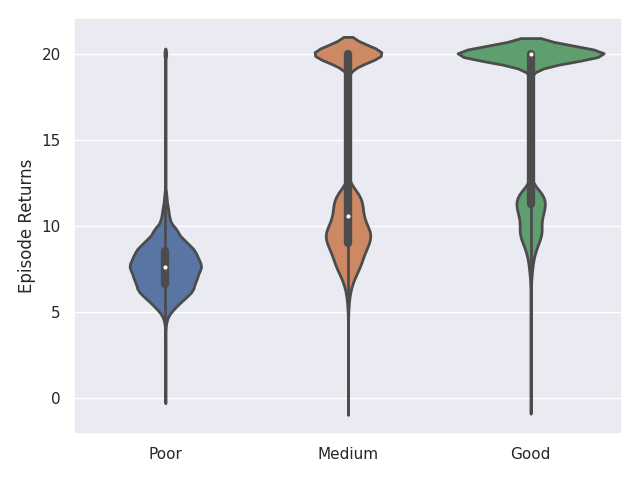}
        \caption{SMAC 5m\_vs\_6m}
        \label{fig:violin_5m_vs_6m_app}
    \end{subfigure}
%    \hfill
    \begin{subfigure}{0.3\textwidth}
        \centering
        \includegraphics[width=\textwidth]{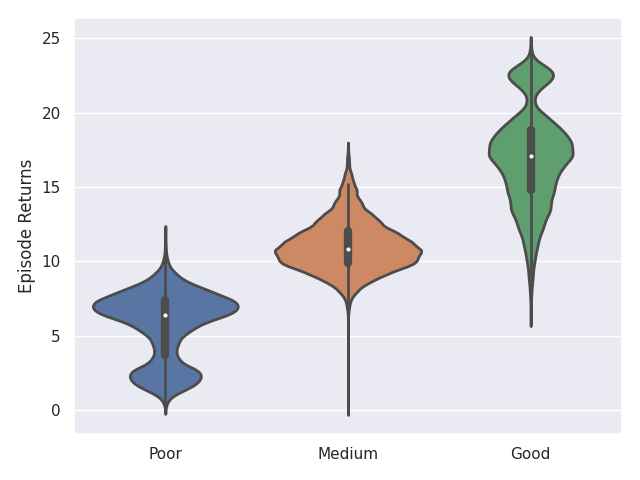}
        \caption{SMAC 3s5z\_vs\_3s6z}
        \label{fig:violin_3s5z_vs_3s6z_app}
    \end{subfigure}
%    \hfill
    \begin{subfigure}{0.3\textwidth}
        \centering
        \includegraphics[width=\textwidth]{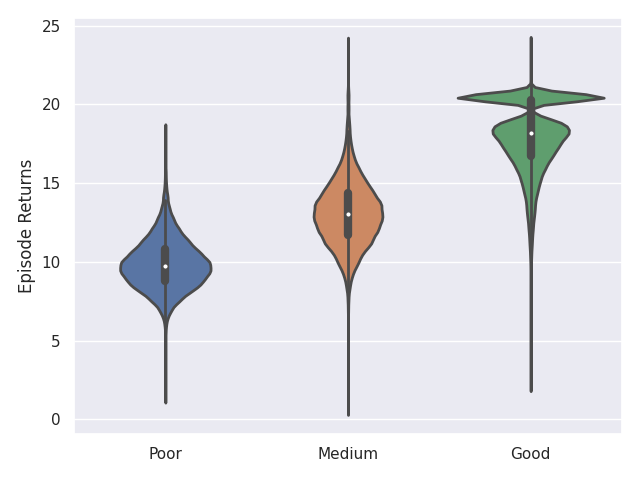}
        \caption{SMAC 2c\_vs\_64zg}
        \label{fig:violin_2c_vs_64zg_app}
    \end{subfigure}
%    \hfill
    \begin{subfigure}{0.3\textwidth}
        \centering
        \includegraphics[width=\textwidth]{figures/violins/27m_vs_30m_all_violin.png}
        \caption{SMAC 27m\_vs\_30m}
        \label{fig:violin_27m_vs_30m_app}
    \end{subfigure}
    %\vfill
    \begin{subfigure}{0.3\textwidth}
       \centering
        \includegraphics[width=\textwidth]{figures/violins/pursuit_all_violin.png}
         \caption{PettingZoo Pursuit}
         \label{fig:violin_pursuit_app}
     \end{subfigure}
     \begin{subfigure}{0.3\textwidth}
         \centering
  \includegraphics[width=\textwidth]{figures/violins/coop_pong_all_violin.png}
         \caption{PettingZoo Cooperative Pong}
         \label{fig:violin_coop_pong_app}
     \end{subfigure}
     %\vfill
     \begin{subfigure}{0.3\textwidth}
         \centering
         \includegraphics[width=\textwidth]{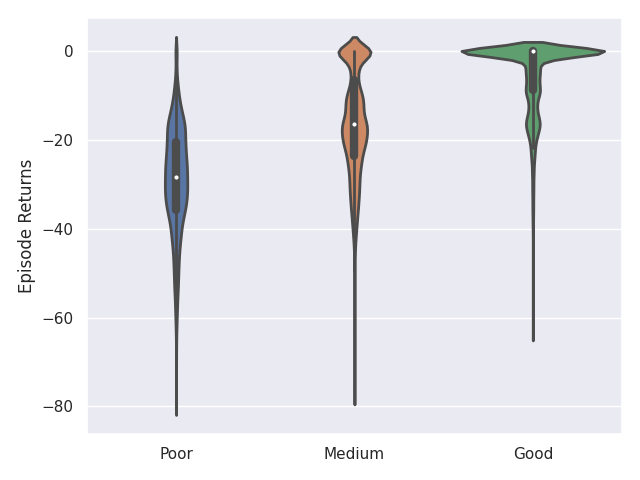}
         \caption{Flatland 3-trains}
         \label{fig:violin_3-train_app}
     \end{subfigure}
     \begin{subfigure}{0.3\textwidth}
         \centering
         \includegraphics[width=\textwidth]{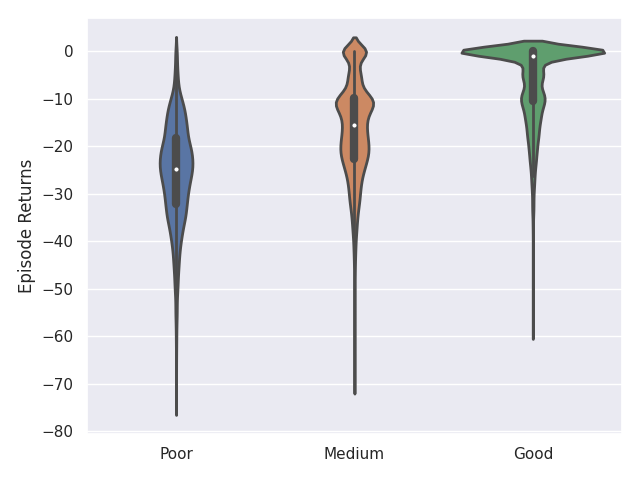}
         \caption{Flatland 5-trains}
         \label{fig:violin_5-train_app}
     \end{subfigure}
    \begin{subfigure}{0.3\textwidth}
         \centering
         \includegraphics[width=\textwidth]{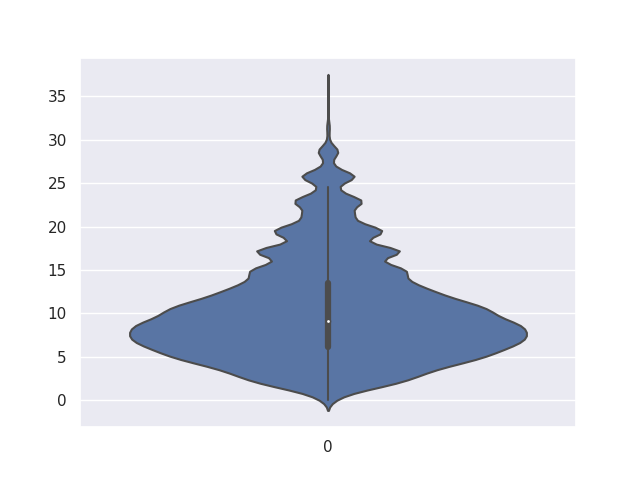}
         \caption{SMAC v2 terran\_5\_vs\_5}
         \label{fig:violin_terran_app}
     \end{subfigure}
    \begin{subfigure}{0.3\textwidth}
         \centering
         \includegraphics[width=\textwidth]{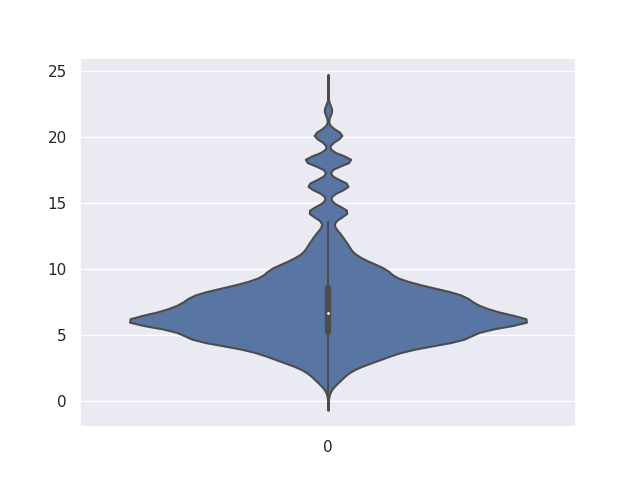}
         \caption{SMAC v2 zerg\_5\_vs\_5}
         \label{fig:violin_zerg_app}
     \end{subfigure}
    \begin{subfigure}{0.3\textwidth}
         \centering
         \includegraphics[width=\textwidth]{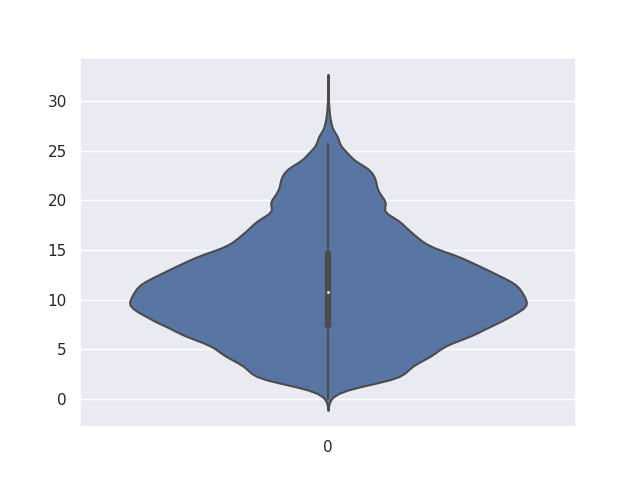}
         \caption{SMAC v2 terran\_10\_vs\_10}
         \label{fig:violin_terran10_app}
     \end{subfigure}
     \caption{Violin plots of all datasets with discrete actions.}
     \label{fig:violin_app}
 \end{figure}

 \begin{figure}[H]
    \centering
    \begin{subfigure}{0.3\textwidth}
        \centering
        \includegraphics[width=\textwidth]{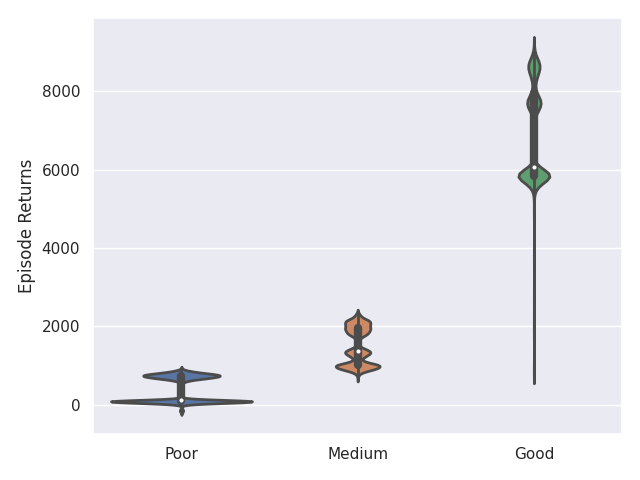}
        \caption{MAMuJoCo 2-halfcheetah}
        \label{fig:violin_2halfcheetah_app}
    \end{subfigure}
    \begin{subfigure}{0.3\textwidth}
        \centering
        \includegraphics[width=\textwidth]{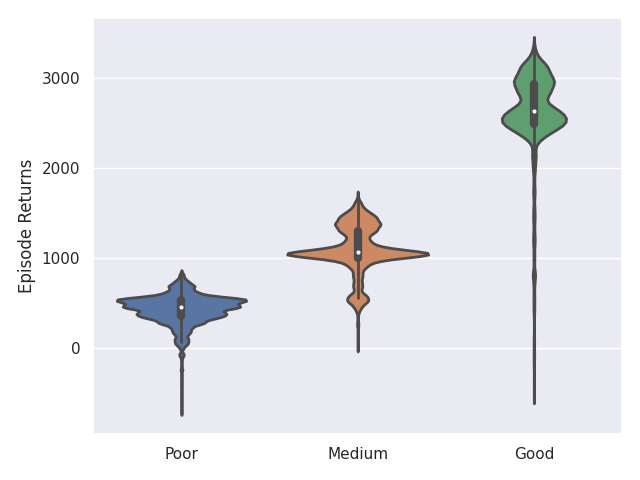}
        \caption{MAMuJoCo 2-ant}
        \label{fig:violin_2ant_app}
    \end{subfigure}
%    \hfill
   \begin{subfigure}{0.3\textwidth}
        \centering
        \includegraphics[width=\textwidth]{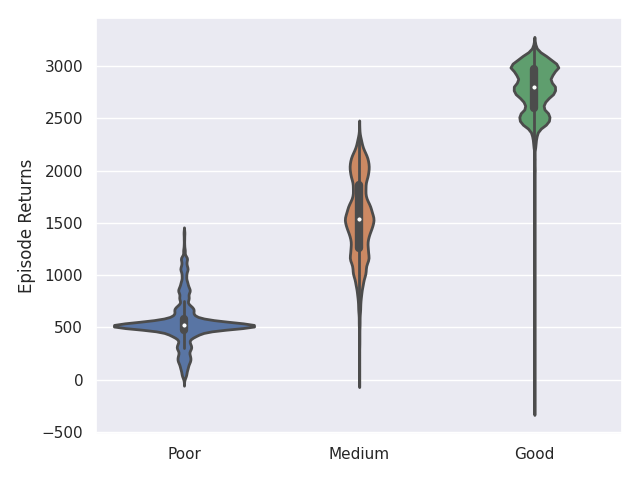}
        \caption{MAMuJoCo 4-ant}
        \label{fig:violin_4ant_app}
    \end{subfigure}
     \begin{subfigure}{0.3\textwidth}
         \centering
         \includegraphics[width=\textwidth]{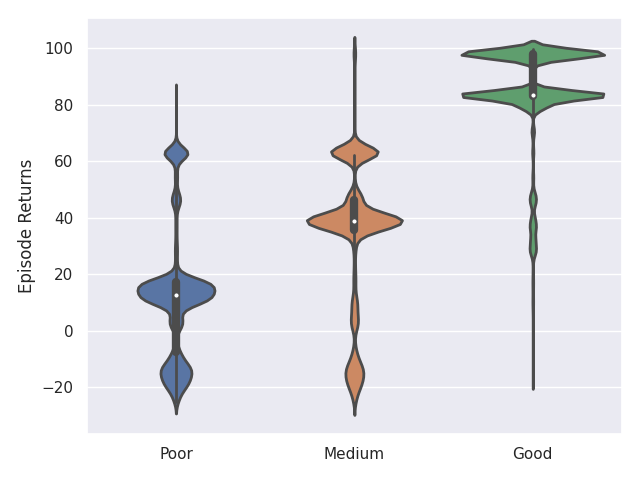}
         \caption{PettingZoo Pistonball}
         \label{fig:violin_pistonball_app}
     \end{subfigure}
    \begin{subfigure}{0.3\textwidth}
        \centering
        \includegraphics[width=\textwidth]{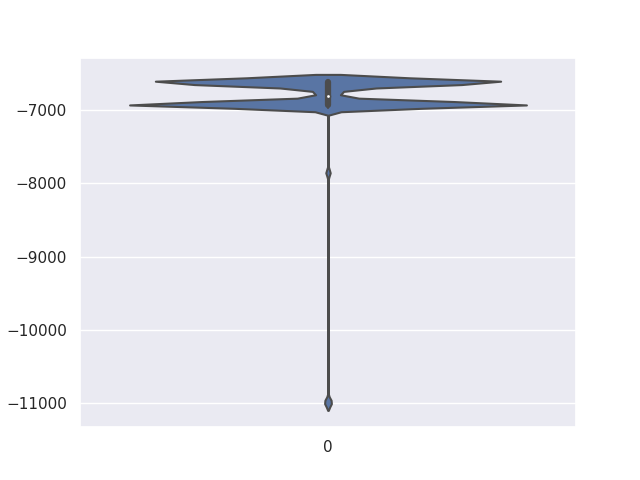}
        \caption{CityLearn}
        \label{fig:violin_citylearn_app}
    \end{subfigure}
    \begin{subfigure}{0.3\textwidth}
        \centering
        \includegraphics[width=\textwidth]{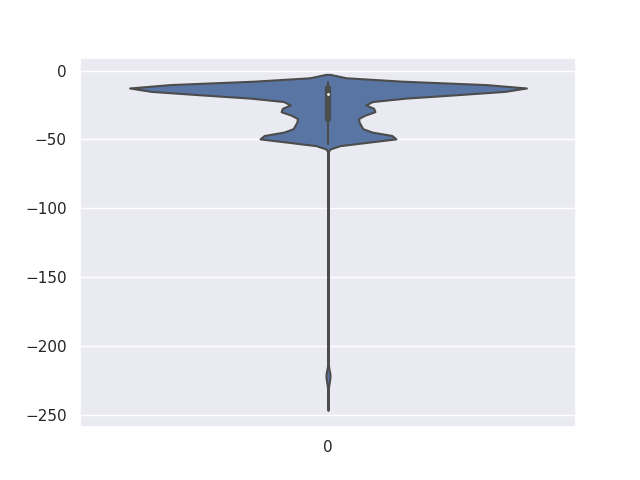}
        \caption{Voltage Control}
        \label{fig:violin_voltage_app}
    \end{subfigure}
     %\vfill
     \caption{Violin plots of all datasets with continuous actions.}
     \label{fig:violin_appendix_cont}
 \end{figure}

\newpage
\section{Additional Baseline Information} \label{sec:appendix_baselines}

In this section, we provide additional implementation details for each of the baseline algorithms implemented in OG-MARL as well as the hyper-parameters used for the experiments and additional results.

\subsection{Background: Single Agent Offline RL Algorithms}

As mentioned in the main text, the primary challenge algorithms need to address during offline training is data distribution mismatch between the behaviour (offline) data and the induced online data. For example, the state visitation frequency induced by the behaviour policy is typically different to that of the learnt policy. While state distribution mismatch can cause failure when the algorithm is deployed, it does not generally cause any issues during training, and can easily be mitigated by expanding the breadth and diversity of the dataset \citep{agarwal2019optimistic}. On the other hand, the most common and difficult-to-address type of distribution mismatch in offline RL is out-of-distribution (OOD) actions. An offline RL algorithm may assign a high value to an OOD action during training due to the extrapolation done by the neural network \citep{fujimoto2019bcq}. These errors then tend to propagate to other state-action pairs, as Q-learning and related algorithms use bootstrapping to compute Bellman targets \citep{Kumar2019StabilizingOQ}. The propagation of extrapolation error then manifests itself as a kind of ``unlearning'', where the performance of the offline RL algorithm rapidly starts to degrade with further training. Most of the remedies proposed in the literature to address OOD actions can be grouped into one of two categories. 

\textbf{Policy constraints.} Several methods try to resitrict the degree to which the learnt policy can become off-policy with respect to the behavioural policy. These methods tend to incorporate some form of behaviour cloning (BC) into RL algorithms to force the learnt policy to remain relatively online with respect to the behaviour dataset. \textit{Batch-Constrained Q-learning} (BCQ)~\citep{fujimoto2019bcq} and \textit{Twin Delayed DDPG + behaviour cloning} (TD3 + BC)~\citep{fujimoto2021td3bc} are two popular algorithms in this class. 

\textbf{Conservative value regularisation.} The second approach mitigates extrapolation error by regularising the learnt value function to avoid overestimating values for OOD actions. An example of this approach, called \textit{conservative Q-learning} (CQL), has been successfully applied to Q-learning and actor-critic methods by \citet{kumar2020cql} in single-agent offline RL.

\subsection{Multi-Agent Offline MARL Algorithms}
At the time of writing, there are only a handful of cooperative offline MARL algorithms available in the literature and we endeavoured to implement as many of them as possible in OG-MARL. However, several algorithms proposed in the literature do not have open-source implementations online and were therefore challenging to re-implement. In \autoref{tab:all_algorithms} we give an overview of all the algorithms in the literature and whether we re-implemented them in OG-MARL. 

\begin{table}[h]
\scriptsize
\centering
\caption{An overview of cooperative offline MARL algorithms from the literature grouped by the work that proposed them as a novel algorithm or baseline. In the second column, we indicate if the code for the algorithm was originally made available online (open-sourced) and in the third column we indicate if the algorithm is implemented in OG-MARL. Algorithms in bold were the main contribution of the respective work while the rest are baselines used in the work. QMIX+CQL and QMIX+BCQ are novel baselines proposed in this work.}
\vspace{1em}
\begin{tabular}{ ccc }
\hline
\textbf{Algorithm Name} & \textbf{Open-Sourced} & \textbf{OG-MARL} \\
\hline
\textbf{MABCQ} & \xmark & \xmark \\
\hline
\textbf{MAICQ} & \cmark & \cmark \\
DOP+CQL & \xmark & \xmark\\
DOP+BCQ & \xmark & \xmark\\
\hline
\textbf{OMAR} & \cmark & \cmark\\
ITD3+CQL & \cmark & \cmark\\
ITD3+BC & \xmark & \cmark \\
MATD3+CQL & \xmark & \cmark \\
MATD3+BC & \xmark & \cmark \\
\hline
QMIX+CQL & n/a & \cmark\\
QMIX+BCQ & n/a & \cmark \\
\hline
\bottomrule
\end{tabular}
\label{tab:all_algorithms}
\end{table}

\subsection{Implementation Details}
In this section, we highlight the most important implementation details for the algorithms in OG-MARL and refer the reader to our open-source code for finer details.

\textbf{QMIX.} Our QMIX implementation is very similar to the original \citep{rashid2018qmix}. We use a single shared Q-network for all agents and concatenate agent IDs to the agent observations so that the network can distinguish between different agents. As in the original QMIX paper, our Q-network is a recurrent network that takes independent agent observations as input, while the mixing network conditions on global state information. To improve the performance of our QMIX implementation, we adopt the recommendation from \citet{hu2021rethinkingQmix} to use \textit{Q-lambda} \citep{sutton1998rlintro} to compute target Q-values.

\textbf{QMIX+CQL.} We add conservative Q-learning \citep{kumar2020cql} to QMIX by uniformly sampling a number of joint-actions from the joint-action space and using those to select Q-values before passing them through the mixing network and using the resulting \textit{mixed} Q-values to calculate the CQL-loss term.

\textbf{QMIX+BCQ.} We add discrete BCQ \citep{fujimoto2019bcq} to QMIX by additionally training a behaviour cloning policy which we use to evaluate how likely each action is to be taken by the behaviour policy given the dataset. If the likelihood is below some threshold, we mask out that action during Q-learning in QMIX.

\textbf{MAICQ.} Our MAICQ implementation is as close as possible to the original by \cite{yang2021believe}.

\textbf{ITD3 and MATD3.} Our ITD3 and MATD3 use a shared policy network and shared Q-network, and concatenate agent IDs to agent observations. The policy is a recurrent neural network with a single GRU layer while the critic is a feedforward neural network that takes the global state as input instead of the observations.

\textbf{ITD3+BC and MATD3+BC.} We incorporate behaviour cloning into ITD3 and MATD3 by adding a behaviour cloning term to the policy learning step as in \citet{fujimoto2021td3bc}

\textbf{ITD3+CQL and MATD3+CQL.} We incorporate conservative Q-learning into ITD3 and MATD3 in a very similar way to how it was done by \cite{pan2022plan}.

\textbf{OMAR} We tried to keep our implementation of OMAR as close to the original \citep{pan2022plan} as possible. The main difference in our implementation is that the policy is a recurrent network, while in the original, they used a feedforward network.

\subsection{Hyper-Parameters}
In this section, we highlight the values we used for the most important hyper-parameters in our benchmark experiments. For additional details about the hyper-parameters we used, we refer to the \texttt{experiments} directory in our open-source code. In \autoref{tab:hyperparams_smac} and \autoref{tab:hyperparams_mamujoco} we give the hyper-parameters for SMAC and MAMuJoCo experiments respectively. In order to keep the online evaluation budget fixed \citep{kurenkov2022onlineEvalBudget}  we tuned hyperparameters on \textit{3m} and \textit{2-Agent HalfCheetah} for SMAC and MAMuJoCo respectively.

\begin{table}[H]
\footnotesize
\centering
\caption{Hyper-Parameters for Discrete Action Algorithms.}
\vspace{1em}
\label{tab:hyperparams_smac}
\begin{tabular}{lll}
Algorithm                          & \textbf{Hyper-Parameter Name}                & \textbf{Value} \\ \hline
\multirow{5}{*}{\textbf{All}}      & Batch Size                                   & 32             \\
                                   & Optimiser                                    & Adam           \\
                                   & Learning Rate                                & 1e-3           \\
                                   & Hidden Activation Function                   & ReLu           \\
                                   & Q-Lambda                                     & 0.6            \\ \hline
\multirow{2}{*}{\textbf{BC}}       & Policy Linear Layer Dimension                & 64             \\
                                   & Policy GRU Layer Dimension                   & 64             \\ \hline
\multirow{5}{*}{\textbf{QMIX}}     & Q-Network Linear Layer Dimension             & 64             \\
                                   & Q-Network Linear Layers Dimension            & 64             \\
                                   & Hyper-Network Dimension                      & 64             \\
                                   & Mixing Embedding Dimension                   & 32             \\
                                   & Soft Target Update Rate                      & 1e-2           \\ \hline
\multirow{4}{*}{\textbf{QMIX+BCQ}} & QMIX Hyper-Parameters                        & Same as above. \\
                                   & Behaviour Network Linear Layer Dimension     & 64             \\
                                   & Behaviour Network GRU Layer Dimension        & 64             \\
                                   & Behaviour Threshold                          & 0.4            \\ \hline
\multirow{3}{*}{\textbf{QMIX+CQL}} & QMIX Hyper-Parameters                        & Same as above. \\
                                   & CQL Alpha                                    & 2.0            \\
                                   & Number of Sampled Actions                    & 20             \\ \hline
\multirow{6}{*}{\textbf{MAICQ}}    & Policy Network Linear Layer Dimension        & 64             \\
                                   & Policy Network GRU Layer Dimension           & 64             \\
                                   & Critic Network First Linear Layer Dimension  & 64             \\
                                   & Critic Network Second Linear Layer Dimension & 64             \\
                                   & Mixing Hyper-Network Dimension               & 64             \\
                                   & Mixing Embedding Dimension                   & 64             \\
                                   & MAICQ Epsilon                                & 0.5            \\
                                   & MAICQ Advantages Beta                        & 0.1            \\
                                   & MAICQ Target Q-Taken Beta                    & 1000           \\ \hline
\bottomrule
\end{tabular}
\end{table}

\begin{table}[H]
\footnotesize
\centering
\caption{Hyper-Parameters for Continuous Action Algorithms.}
\vspace{1em}
\label{tab:hyperparams_mamujoco}
\begin{tabular}{lll}
Algorithm                          & \textbf{Hyper-Parameter Name}                & \textbf{Value} \\ \hline
\multirow{5}{*}{\textbf{All}}      & Batch Size                                   & 32             \\
                                   & Optimiser                                    & Adam           \\
                                   & Learning Rate                                & 5e-4           \\
                                   & Hidden Activation Function                   & ReLu           \\
                                   & Policy Linear Layer Dimension                & 128             \\
                                   & Policy GRU Layer Dimension                   & 128             \\ \hline
\multirow{3}{*}{\textbf{ITD3}}     & Critic Linear Layer Dimension             & 128             \\
                                   &Critic Linear Layers Dimension            & 128             \\
                                   & Target Update Rate & 0.01 \\ \hline
\multirow{1}{*}{\textbf{ITD3+BC}}  & Behaviour Cloning Alpha                          & 2.5     \\ \hline
\multirow{2}{*}{\textbf{ITD3+CQL}} & CQL Alpha                                    & 10.0            \\
                                   & Number of OOD Actions                    & 10             \\ \hline
\multirow{6}{*}{\textbf{OMAR}}    & CQL Parameters        & Same as above.             \\
                                   & OMAR Iterations                                & 3           \\
                                   & OMAR Number of Samples                        & 20            \\
                                   & OMAR Number of Elites                    & 5           \\
                                   & OMAR Sigma                    & 2.0           \\
                                   & OMAR Coefficient                   & 0.7           \\ \hline

\bottomrule
\end{tabular}
\end{table}

\newpage
\subsection{Additional Results}
In this section, we provide additional baseline results on datasets in OG-MARL.

\begin{figure}[H]
    \begin{subfigure}{0.33\textwidth}
        \includegraphics[width=\textwidth]{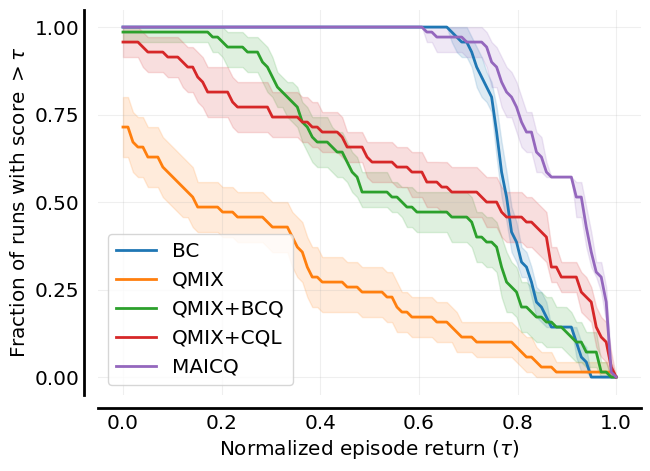}
        \caption{SMAC Good}
    \end{subfigure}
    \begin{subfigure}{0.33\textwidth}
        \includegraphics[width=\textwidth]{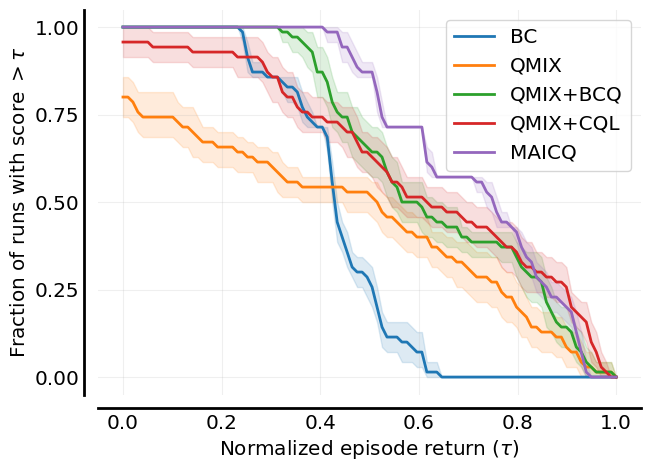}
        \caption{SMAC Medium}
    \end{subfigure}
    \begin{subfigure}{0.33\textwidth}
        \includegraphics[width=\textwidth]{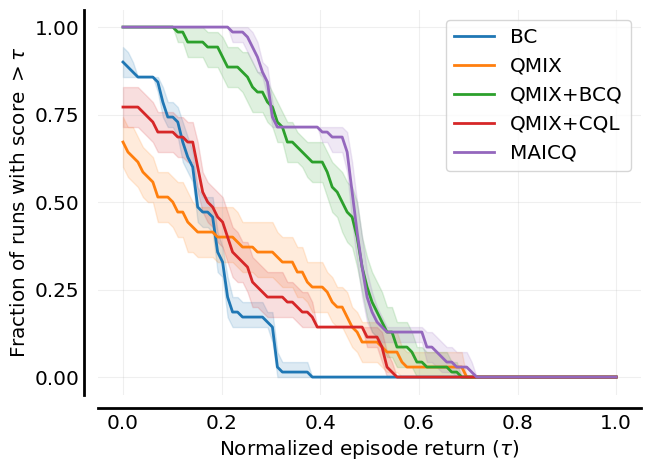}
        \caption{SMAC Poor}
    \end{subfigure}
    \caption{Aggregated performance profiles \citep{agarwal2021deep} for SMAC. Shaded regions show pointwise 95\% confidence bands based on percentile bootstrap with stratified sampling. Results were aggregated across all scenarios and seeds.}
    \label{fig:smac_perf_profiles}
\end{figure}

\textbf{Discrete Actions.} In \autoref{tab:all_discrete_results} we give the baseline results on datasets with discrete actions. In \autoref{fig:smac_perf_profiles} we provide the aggregated performance profiles for SMAC.

\begin{table}[H]
\scriptsize
\centering
\caption{Baseline results on datasets with discrete actions. The mean episode return with one standard deviation across all seeds is given. The best mean episode return in each row is given in bold.}
\vspace{1em}
\begin{tabular}{llllllll}
\textbf{Environment}             & \multicolumn{1}{c}{\textbf{Scenario}} & \multicolumn{1}{c}{\textbf{Dataset}} & \multicolumn{1}{c}{\textbf{BC}} & \multicolumn{1}{c}{\textbf{QMIX}} & \multicolumn{1}{c}{\textbf{QMIX+BCQ}} & \multicolumn{1}{c}{\textbf{QMIX+CQL}} & \multicolumn{1}{c}{\textbf{MAICQ}} \\ \hline
\multirow{21}{*}{SMAC} & \multirow{3}{*}{3m}                   & Good             & 16.0$\pm$1.0                    & 13.8$\pm$4.5                      & 16.3$\pm$1.5                          & \textbf{19.6$\pm$0.3}                 & 18.8$\pm$0.6                       \\
&                                      & Medium           & 8.2$\pm$0.8                     & 17.3$\pm$0.9                      & 18.3$\pm$1.2                          & \textbf{18.9$\pm$1.2}                 & 18.1$\pm$0.7                       \\
&                                      & Poor             & 4.4$\pm$0.1                     & 10.0$\pm$2.9                      & 12.4$\pm$2.3                          & 5.8$\pm$0.4                           & \textbf{14.4$\pm$1.2}              \\ \cline{2-8}
& \multirow{3}{*}{8m}                   & Good             & 16.7$\pm$0.4                    & 4.6$\pm$2.8                       & 12.7$\pm$6.3                          & 11.3$\pm$6.1                          & \textbf{19.6$\pm$0.3}              \\
&                                       & Medium           & 10.7$\pm$05                     & 13.9$\pm$1.6                      & 16.0$\pm$1.4                          & 16.8$\pm$3.1                          & \textbf{18.6$\pm$0.5}              \\
&                                       & Poor             & 5.3$\pm$0.1                     & 6.0$\pm$1.3                       & 5.8$\pm$1.4                           & 4.6$\pm$2.4                           & \textbf{10.8$\pm$0.8}              \\ \cline{2-8}
& \multirow{3}{*}{5m\_vs\_6m}           & Good             & \textbf{16.6$\pm$0.6}           & 8.0$\pm$0.5                       & 8.3$\pm$0.9                           & 13.8$\pm$3.9                          & 16.3$\pm$0.9                       \\
&                                      & Medium           & 12.4$\pm$0.9                    & 11.9$\pm$1.1                      & 12.1$\pm$1.3                          & \textbf{16.9$\pm$1.2}                 & 15.3$\pm$0.7                       \\
&                                      & Poor             & 7.5$\pm$0.2                     & 10.7$\pm$0.9                      & \textbf{11.0$\pm$0.9}                 & 10.4$\pm$1.0                          & 9.4$\pm$0.4                        \\ \cline{2-8}
& \multirow{3}{*}{27m\_vs\_30m}         & Good             & 15.7$\pm$0.3                    & 3.2$\pm$1.4                       & 10.2$\pm$1.4                          & 6.0$\pm$3.3                           & \textbf{16.1$\pm$1.8}              \\
&                                      & Medium           & 10.3$\pm$0.4                    & 6.2$\pm$2.1                       & 9.8$\pm$1.2                           & 8.0$\pm$1.7                           & \textbf{12.9$\pm$0.5}              \\
&                                      & Poor             & 6.0$\pm$1.5                     & 2.1$\pm$1.7                       & \textbf{10.3$\pm$0.7}                 & 3.7$\pm$2.7                           & 10.1$\pm$0.8                       \\ \cline{2-8}
& \multirow{3}{*}{2s3z}                 & Good             & 18.2$\pm$0.4                    & 5.9$\pm$3.4                       & 16.6$\pm$1.2                          & 19.0$\pm$0.8                          & \textbf{19.6$\pm$0.3}              \\
&                                       & Medium           & 12.3$\pm$0.7                    & 5.2$\pm$0.9                       & 13.6$\pm$1.5                          & 14.3$\pm$2.0                          & \textbf{17.2$\pm$0.6}              \\
&                                       & Poor             & 6.7$\pm$0.3                     & 3.8$\pm$1.2                       & 11.5$\pm$1.0                          & 10.1$\pm$0.7                          & \textbf{12.1$\pm$0.4}              \\ \cline{2-8}
& \multirow{3}{*}{3s5z\_vs\_3s6z}       & Good             & 15.0$\pm$0.6                    & 3.1$\pm$1.3                       & 8.4$\pm$0.7                           & 7.3$\pm$1.9                           & \textbf{16.2$\pm$0.7}              \\
&                                       & Medium           & 10.6$\pm$0.2                    & 3.0$\pm$1.0                       & 10.5$\pm$0.8                          & 8.1$\pm$3.1                           & \textbf{12.3$\pm$0.3}              \\
&                                      & Poor             & 6.1$\pm$0.3                     & 2.8$\pm$1.0                       & 8.2$\pm$0.9                           & 2.9$\pm$0.9                           & \textbf{8.4$\pm$0.2}               \\ \cline{2-8}
& \multirow{3}{*}{2c\_vs\_64zg}         & Good             & 17.5$\pm$0.4                    & 10.9$\pm$4.0                      & 18.7$\pm$0.8                          & 18.1$\pm$0.8                          & \textbf{19.3$\pm$0.3}              \\
&                                       & Medium           & 12.5$\pm$0.3                    & 16.8$\pm$1.6                      & \textbf{18.4$\pm$0.5}                 & 14.9$\pm$0.7                          & 14.6$\pm$0.6                       \\
&                                      & Poor             & 9.7$\pm$0.2                     & 11.6$\pm$2.2                      & \textbf{14.3$\pm$0.8}                 & 12.1$\pm$0.4                          & 12.5$\pm$0.4                       \\ \hline
\multirow{6}{*}{PettingZoo}   &   \multirow{3}{*}{Co-op Pong}    & Good             & 31.2$\pm$3.5                    & 0.6$\pm$3.5                       & 1.9$\pm$1.1                           & \textbf{90.0$\pm$4.7}                 & 75.4$\pm$3.9                       \\
&                          & Medium           & 21.6$\pm$4.8                    & 10.6$\pm$17.6                     & 20.3$\pm$12.2                         & 64.9$\pm$15.0                         & \textbf{84.6$\pm$0.9}              \\
&                          & Poor             & 1.0$\pm$0.9                     & 14.4$\pm$16.0                     & 30.2$\pm$20.7                         & 52.7$\pm$8.5                          & \textbf{74.8$\pm$7.8}              \\ \cline{2-8}
& \multirow{3}{*}{Pursuit} & Good             & 78.3$\pm$1.8                    & 6.7$\pm$19.0                      & 66.9$\pm$14.0                         & 54.4$\pm$6.3                          & \textbf{92.7$\pm$3.7}              \\
&                          & Medium           & 15.0$\pm$1.6                    & -24.4$\pm$20.2                    & 16.6$\pm$10.7                         & 20.6$\pm$10.3                         & \textbf{35.3$\pm$3.0}              \\
&                          & Poor             & -18.5$\pm$1.6                   & -43.7$\pm$5.6                     & \textbf{-0.7$\pm$4.0}                 & -19.6$\pm$3.3                         & -4.1$\pm$0.7                       \\ \hline 
\multirow{6}{*}{Flatland}        & \multirow{3}{*}{3 Trains}             & Good                                 & -5.6$\pm$2.4                     & -3.6$\pm$0.4                       & -3.5$\pm$2.8                           & \textbf{-2.1$\pm$0.4}                           & -25.3$\pm$0.2                         \\
                                 &                                       & Medium                               & \textbf{-4.5$\pm$2.5}                    & -12.5$\pm$1.0                       & -7.1$\pm$2.7                           & -4.6$\pm$0.5                           & -25.2$\pm$0.2                         \\
                                 &                                       & Poor                                 & \textbf{-11.4$\pm$3.8}                   & -27.9$\pm$0.8                      & -17.3$\pm$4.1                           & -24.9$\pm$0.4                           & -25.9$\pm$0.9                         \\ \cline{2-8} 
                                 & \multirow{3}{*}{5 Trains}             & Good                                 & -28.1$\pm$1.4                    & -6.4$\pm$0.6                       & -8.1$\pm$4.9                           & \textbf{-3.2$\pm$0.5}                           & -25.6$\pm$0.5                        \\
                                 &                                       & Medium                               & -9.7$\pm$5.7                     & -17.9$\pm$1.0                       & -10.6$\pm$8.4                           & \textbf{-3.7$\pm$0.3}                           & -25.9$\pm$0.5                        \\
                                 &                                       & Poor                                 & -9.9$\pm$3.8                     & -24.7$\pm$1.9                       & \textbf{-9.5$\pm$6.6}                          & -11.1$\pm$4.0                           & -25.6$\pm$0.4                       \\ \hline
\multirow{3}{*}{SMACv2}        & \multirow{1}{*}{terran\_5\_vs\_5}             & Replay                                 & 7.3$\pm$1.0                     & 13.7$\pm$2.7                       & \textbf{13.8$\pm$4.4}                           & 11.8$\pm$0.9                           & 13.7$\pm$1.7  \\ \cline{2-8} 
                                & \multirow{1}{*}{zerg\_5\_vs\_5}             & Replay                                 & 6.8$\pm$0.6                    & 10.2$\pm$2.4                       & 10.3$\pm$1.2                           & 10.3$\pm$3.4                           & \textbf{10.6$\pm$0.7}   \\ \cline{2-8} 
                                & \multirow{1}{*}{terran\_10\_vs\_10}             & Replay                                 & 7.4$\pm$0.5                    & 10.4$\pm$2.5                       & 12.7$\pm$2.0                           & 11.8$\pm$2.0                           & \textbf{14.4$\pm$0.7}   \\ \hline
\bottomrule
\end{tabular}
\label{tab:all_discrete_results}
\end{table}

\begin{figure}[H]
    \begin{subfigure}{0.33\textwidth}
        \includegraphics[width=\textwidth]{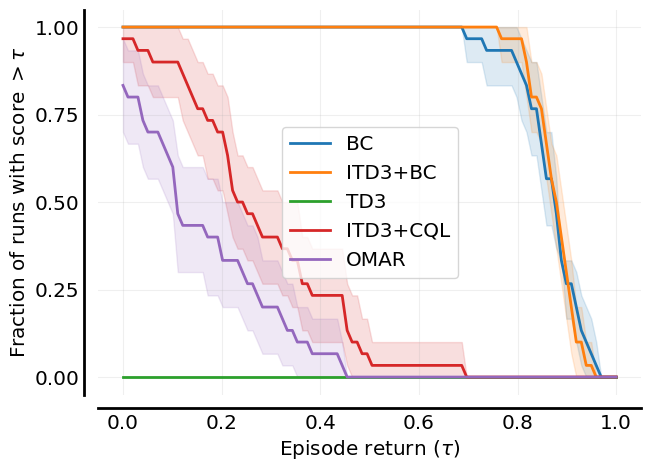}
        \caption{MAMuJoCo Good}
    \end{subfigure}
    \begin{subfigure}{0.33\textwidth}
        \includegraphics[width=\textwidth]{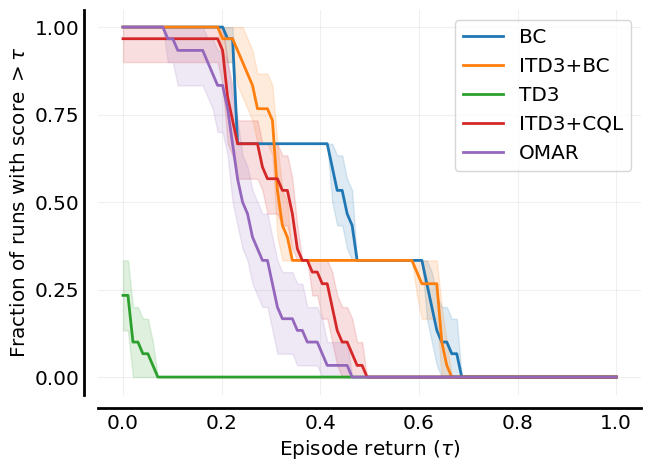}
        \caption{MAMuJoCo Medium}
    \end{subfigure}
    \begin{subfigure}{0.33\textwidth}
        \includegraphics[width=\textwidth]{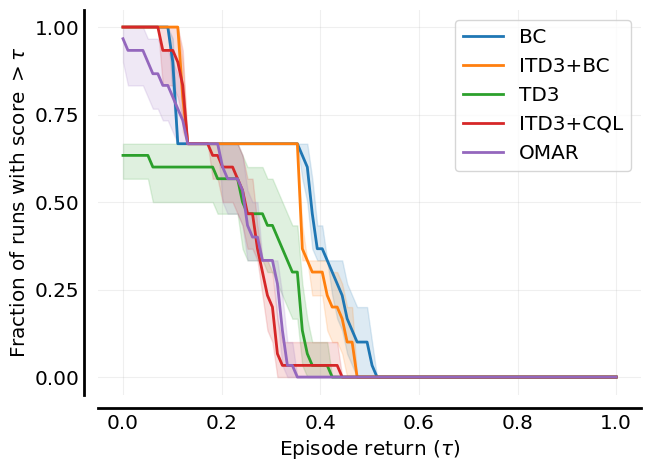}
        \caption{MAMuJoCo Poor}
    \end{subfigure}
    \caption{Aggregated performance profiles \citep{agarwal2021deep} for MAMuJoCo. Shaded regions show pointwise 95\% confidence bands based on percentile bootstrap with stratified sampling. Results were aggregated across all scenarios and seeds.}
    \label{fig:mamujoco_perf_profiles}
\end{figure}

\newpage
\textbf{Continuous Actions.} In \autoref{tab:all_cont_results} we give the baseline results on datasets with continuous actions. In \autoref{fig:mamujoco_perf_profiles} we provide the aggregated performance profiles for MAMuJoCo.

% \usepackage{multirow}
\begin{table}[H]
\scriptsize
\centering
\caption{Baseline results on datasets with continuous actions. The mean episode return with one standard deviation across all seeds is given. The best mean episode return in each row is given in bold.}
\vspace{1em}
\begin{tabular}{llclllll}
\textbf{Environment} & \textbf{Scenario}              & \multicolumn{1}{c}{\textbf{Dataset}} & \multicolumn{1}{c}{\textbf{BC}} & \multicolumn{1}{c}{\textbf{ITD3}} & \multicolumn{1}{c}{\textbf{ITD3+BC}} & \multicolumn{1}{c}{\textbf{ITD3+CQL}} & \multicolumn{1}{c}{\textbf{OMAR}} \\ \hline
\multirow{9}{*}{MAMuJoCo} & \multirow{3}{*}{2-HalfCheetah} & Good                                 & 6846$\pm$574                    & -578$\pm$33                       & \textbf{7025$\pm$439}                & 2934$\pm$1666                         & 1434$\pm$1903                     \\
                             &  & Medium                               & 1627$\pm$187                    & -87$\pm$223                       & \textbf{2561$\pm$82}                 & 1755$\pm$283                          & 1892$\pm$220                      \\
                             &  & Poor                                 & 465$\pm$59                  & -392$\pm$76                       & 736$\pm$72                           & \textbf{739$\pm$191}                  & 384$\pm$420                       \\ \cline{2-8} 
& \multirow{3}{*}{2-Ant}         & Good                                 & 2697$\pm$267                    & -1274$\pm$501                     & \textbf{2922$\pm$194}                & 606$\pm$487                           & 464$\pm$469                       \\
                             &  & Medium                               & \textbf{1145$\pm$126}           & -1416$\pm$845                     & 744$\pm$283                          & 716$\pm$431                           & 799$\pm$186                       \\
                             &  & Poor                                 & 954$\pm$80                      & 741$\pm$398                       & \textbf{1256$\pm$122}                & 814$\pm$177                           & 857$\pm$73                        \\ \cline{2-8}
& \multirow{3}{*}{4-Ant}        & Good                                 & \textbf{2802$\pm$133}           & -1033$\pm$432                     & 2628$\pm$971                         & 712$\pm$672                           & 344$\pm$631                       \\
                              & & Medium                               & 1617$\pm$153                    & -1159$\pm$733                     & \textbf{1843$\pm$494}                & 1190$\pm$186                          & 929$\pm$349                       \\
                              & & Poor                                 & 1033$\pm$122                    & 703$\pm$465                       & \textbf{1075$\pm$96}                 & 518$\pm$122                           & 518$\pm$112                       \\ \hline

\multirow{3}{*}{PettingZoo}      & \multirow{3}{*}{Pistonball}           & \multicolumn{1}{l}{Good}   & \textbf{94.1$\pm$1.2}                     & 0.8$\pm$10.6                       & 93.1$\pm$2.0                           & -\tablefootnote{Due to the nature of the CQL and OMAR algorithms, and the large number of agents in Pistonball we have not managed to successfully run these experiments without running out of RAM on the compute available to us. We are working on resolving this.}                          & -                        \\
                                     &                                       &   \multicolumn{1}{l}{Medium} & 10.3$\pm$6.0                     & -5.5$\pm$3.2                       & \textbf{13.7$\pm$8.9 }                          & -                           & -                        \\
                                 &                                       & \multicolumn{1}{l}{Poor}   & 4.6$\pm$3.2                     & -5.8$\pm$2.3                       & \textbf{6.1$\pm$2.5}                           & -                           & -                        \\ \hline
\multirow{1}{*}{CityLearn}      & \multirow{1}{*}{2022\_all\_phases}           & \multicolumn{1}{l}{Replay}   & \textbf{-6576$\pm$39}                     & -6594$\pm$1                       & -6663$\pm$87                           & -6598$\pm$9                         &  -6630$\pm$44                       \\ \hline
\multirow{1}{*}{VoltageControl}      & \multirow{1}{*}{case33\_3min\_final}           & \multicolumn{1}{l}{Replay}   & \textbf{-9.9$\pm$2.3}                     & -10.0$\pm$0.8                       & -11.1$\pm$0.7                          & -32.9$\pm$6.7                        & -26.5$\pm$9.4             \\ \hline
\bottomrule
\end{tabular}
\label{tab:all_cont_results}
\end{table}

\subsection{Reproducing Baseline Results}
Scripts for reproducing our baseline experiments are included in the open-sourced code.

\subsection{Baseline Compute Budget}
To run all of our baselines we used CPUs on an internal compute cluster. In total we used 546 days of CPU compute time.

\newpage
\section{Dataset Licence, Author Statement, Hosting \& Maintenance Plan}\label{sec:licence}

\subsection{Dataset Licence}
The datasets in OG-MARL are licenced under the Common Dataset Licences, \textbf{CC BY-NC-SA}.\footnote{\url{https://paperswithcode.com/datasets/license}} This license allows reusers to distribute, remix, adapt, and build upon the material in any medium or format for noncommercial purposes only, and only so long as attribution is given to the creator. If you remix, adapt, or build upon the material, you must license the modified material under identical terms.

\subsection{Author Statement}
The authors of "Off-the-Grid MARL: Datasets with Baselines for Offline Multi-Agent Reinforcement Learning" bear all responsibility in case of any violation of rights during the collection of the data or other work, and will take appropriate action when needed, e.g. to remove data with such issues.

\subsection{Hosting \& Maintenance Plan}

The OG-MARL datasets are hosted in an accessible, online storage bucket, kindly hosted by InstaDeep. An easy-to-use interface for downloading datasets from the bucket is provided via our website. Datasets will continue to be maintained by the authors and dataset versions will be tracked on the OG-MARL GitHub repository. The OG-MARL code will also be tracked on GitHub. Issues and feature requests can be submitted on the GitHub repository. 

\end{document}